\documentclass{article} 
\usepackage{times}
\usepackage{graphicx} 
\usepackage{subcaption} 

\usepackage{natbib}

\usepackage{algorithm}
\usepackage{algorithmic}

\usepackage{hyperref}

\usepackage[final]{nips_2017}
\usepackage{url}
\usepackage{bm}
\usepackage{mathtools}
\usepackage{amsmath}
\usepackage{amsfonts}
\usepackage{cleveref}
\usepackage{graphicx}
\usepackage{dirtytalk}
\usepackage{pbox}
\usepackage{cleveref}

\title{Fast Adaptation in Generative Models with Generative Matching Networks}

\author{
  Sergey Bartunov \\
  Higher School of Economics \\
  \texttt{sbos@sbos.in} \\
   \And
   Dmitry Vetrov \\
   Yandex, Higher School of Economics \\
   \texttt{vetrovd@yandex.ru} \\
}

\begin{document}

\newcommand{\obsvar}{x}
\newcommand{\hidvar}{z}
\newcommand{\query}{\mathbf{q}}
\newcommand{\hstate}{h}
\newcommand{\result}{r}
\newcommand{\obsvars}{\mathbf{x}}
\newcommand{\hidvars}{\mathbf{z}}
\newcommand{\Obsvars}{\mathbf{X}}
\newcommand{\Hidvars}{\mathbf{Z}}
\newcommand{\globvars}{\boldsymbol{\alpha}}
\newcommand{\noisevars}{\boldsymbol{\epsilon}}
\newcommand{\param}{\theta}
\newcommand{\params}{\boldsymbol{\theta}}
\newcommand{\approxfam}{\mathcal{Q}}
\newcommand{\approxfamhid}{\mathcal{Q}_{\hidvars}}
\newcommand{\approxfamhidd}[1]{\mathcal{Q}_{\hidvar_{#1}}}
\newcommand{\approxfampar}{\mathcal{Q}_{\params}}
\newcommand{\varparam}{\phi}
\newcommand{\varparams}{\boldsymbol{\phi}}
\newcommand{\glparams}{\boldsymbol{\pha}}

\maketitle

\begin{abstract} 
Despite recent advances, the remaining bottlenecks in deep generative models are necessity of extensive training and difficulties with generalization from small number of training examples.
We develop a new generative model called Generative Matching Network which is inspired by the recently proposed matching networks for one-shot learning in discriminative tasks.
By conditioning on the additional input dataset, our model can  instantly learn new concepts that were not available in the training data but conform to a similar generative process.
The proposed framework does not explicitly restrict diversity of the conditioning data and also does not require an extensive inference procedure for training or adaptation.
Our experiments on the Omniglot dataset demonstrate that Generative Matching Networks significantly improve predictive performance on the fly as more additional data is available and outperform existing state of the art conditional generative models.
\end{abstract} 

\section{Introduction}

Deep generative models are currently one of the most promising directions in generative modelling. 
In this class of models the generative process is defined by a composition of conditional distributions modelled using deep neural networks which form a hierarchy of latent and observed variables.
This approach allows to build models with complex, non-linear dependencies between variables and efficiently learn the variability across training examples.

Such models are trained by stochastic gradient methods which can handle large datasets and a wide variety of model architectures but also present certain limitations.
The training process usually consists of small, incremental updates of networks' parameters and requires many passes over training data.
Notably, once a model is trained, it cannot be adapted to newly available data without complete re-training to avoid catastrophic interference~\citep{mccloskey1989catastrophic, ratcliff1990connectionist}. 
There is also a risk of overfitting for concepts that are not represented by enough training examples which is caused by high capacity of the models.
Hence, most of deep generative models are not well-suited for rapid learning in real-world applications where data acquisition is expensive or fast adaptation to new data is required.


We present Generative Matching Network (GMN)\footnote{Source code is available at \url{https://github.com/sbos/gmn}}, a deep generative model suitable for fast learning in the few-shot setting, that makes progress in both of these directions.
After the model is trained on a particular domain, it can instantly adapt it's generative distribution by conditioning on the additional dataset from a similar domain, without invoking a computationally extensive inference procedure.
Generative matching network is inspired the attentional mechanism implemented in Matching Networks, originally proposed for supervised discriminative tasks~\citep{vinyals2016matching}.
The attentional mechanism, extended to unsupervised learning tasks, allows to smoothly interpolate between similar examples in the conditioning data and implicitly account for the underlying class structure, hence being robust to interference from diverse data, which was not previously demonstrated by analogous models.



\section{Background}\label{sec:background}

We consider the problem of learning a probabilistic generative model which can be expressed as a probability distribution $p(\obsvars | \params)$ over objects of interests $\obsvars$ parameterized by $\params$. 
The major class of generative models introduce also \emph{latent} variables $\hidvars$ that are used to explain or generate an object $\obsvars$ such that $p(\obsvars | \params) = \int p(\hidvars | \params) p(\obsvars | \hidvars, \params) d \hidvars$ and assumed to be non-observable.

Currently, the common practice is to restrict the conditional distributions $p(\hidvars | \params)$ and $p(\obsvars | \hidvars, \params)$ to tractable distribution families and use deep neural networks for regressing their parameters. 
The expressive power of deep non-linear generative models comes at a price since neither marginal distribution $p(\obsvars | \params)$ can be computed analytically nor it can be directly optimized in a statistically efficient way.
Fortunately, intractable maximum likelihood training can be avoided in practice by resorting to adversarial training~\citep{gutmann2012noise, goodfellow2014generative} or variational inference framework~\citep{kingma2013auto, rezende2014stochastic} which we consider further. 

\subsection{Training with variational inference}\label{sec:var_inf}

Recent developments in variational inference alleviate problems with maximizing the intractable marginal likelihood $\log p(\obsvars | \params)$ by approximating it with a lower bound~\citep{jordan1999introduction}: 
\begin{equation}\label{eq:elbo}
    \log p(\obsvars | \params) \geq \mathcal{L}(\params, \varparams) =  \mathbb{E}_{q} \left[ \log p(\obsvars, \hidvars | \params) - \log q(\hidvars | \obsvars, \varparams) \right] 
\end{equation}
Tightness of the bound is controlled by the recognition model $q(\hidvars | \obsvars, \varparams)$ which aims to minimize Kullback-Leibler divergence to the true posterior $p(\hidvars | \obsvars, \params)$. 

Similarly to the generative model, recognition model may also be implemented with the use of deep neural networks or other parameter regression which is known as \emph{amortized inference}~\citep{gershman2014amortized}.
Amortized inference allows to use a single recognition model for many training examples.
Thus, it is convenient to perform training of the generative model $p(\obsvars | \params)$ by stochastic gradient optimization of variational lower bounds~\eqref{eq:elbo} corresponding to independent observations $\{ \obsvars_i \}_{i=1}^N$.

The clear advantage of this approach is its scalability.
Every stochastic update to the parameters computed from only a small portion of training examples has an immediate effect on the whole dataset.
However, while a single parameter update may be relatively fast, a large number of them is required to significantly improve generative or inferential performance of the model.
Hence, gradient training of generative models usually results into an extensive computational process which prevents from rapid incremental learning.
In the next section we discuss potential solutions to this problem that allow to implement fast learning ability in generative models.

\subsection{Fast learning in generative models}\label{sec:adaptation}

In probabilistic modelling framework the natural way of incorporating knowledge about newly available data is conditioning. 
One may design a model that being conditioned on the additional input data $\Obsvars = \obsvars_1, \obsvars_2, \ldots, \obsvars_T$ represents a new generative distribution $p(\obsvars | \Obsvars, \params)$.

An implementation of this idea can be found in the model by~\citet{rezende2016one} which was able to produce new examples of a concept that was missing at the training time but had similarities in the underlying generative process with the other training examples. 
The model supported an explicit conditioning on a single observation $\obsvars'$ representing the new concept to construct a new generative distribution of the form $p(\obsvars | \obsvars', \params)$.

The explicit conditioning when adaptation is performed \emph{by the model} itself and and has to be learned is not the only way to propagate knowledge about new data. 
Another solution which is often encountered in Bayesian models is to maintain a \emph{global} latent variable encoding information about the whole available dataset such that the individual observations are conditionally independent given it's value. 
Denoting the global variable as $\globvars$, a typical model from this class would have the following form:
The model then would have the following form: $p(\Obsvars | \params) = \int p(\globvars | \params) \prod_{t=1}^T p(\obsvars_t | \globvars, \params) d \globvars$.

The principal existence of such a global variable may be justified by the de Finetti's theorem~\citep{diaconis1980finite} under the exchangeability assumption.
In global latent variable models, the conditional generative distribution $p(\obsvars | \Obsvars, \params)$ is then defined implicitly via posterior over the global variable:
$
    p(\obsvars | \Obsvars, \params) = \int p(\obsvars | \globvars, \params) p(\globvars | \Obsvars, \params) d \globvars. $
Once there is an efficient inference procedure for the global variable $\globvars$, learning or fast adaptation can be implemented straightforwardly. 

There are several relevant examples of generative models with global latent variables used for model adaptation and few-shot learning. 
\citet{salakhutdinov2013learning} combined deep Boltzmann machine (DBM) with nested Dirichlet process (nDP) in a Hierarchical-Deep (HD) model. 
While DBM was used to learn low-level features, the nonparametric distribution over high-level features defined via nDP allowed to infer a latent global hierarchy of concepts from the training data.
Later,~\citet{lake2015human} proposed Bayesian program learning (BPL) approach for building a generative model of handwritten characters.
The model was defined as a probabilistic program contained fine-grained specification of prior knowledge of the task such as generation of strokes and their composition into characters mimicking human drawing behaviour. 

While being suitable for learning from small data, both HD and BPL models required extensive sampling as a necessary part of either training or generation procedures.
Hence, although Bayesian inference over the global latent variable may prevent overfitting, \emph{fast} learning still remains a challenge for sampling-based inference.

The recently proposed neural statistician model~\citep{edwards2016towards} is another deep generative model with a global latent variable. 
The model was trained by optimizing a variational lower bound following the approach described in section~\ref{sec:var_inf}, but with an additional recognition model approximating posterior distribution over the global latent variable.
Authors designed the recognition model to be computationally efficient and require only a single pass over data which consisted of extracting special features from the examples, applying to them a pooling operation (e.g. averaging) and passing the result to another network providing parameters of the variational approximation.

This simple architecture allowed for the fast learning and guaranteed invariance to both data permutations and size of the conditioning dataset.
However, authors evaluated the fast learning ability in the model  only in the setting where all of the training examples represented the same single concept.
Indeed, as we show later in section~\ref{sec:few_shot}, this approach is less efficient for adaptation to more complex data, perhaps because a fixed parametric description is too restrictive for an accurate representation of datasets of varying complexity.

\section{Generative Matching Networks}\label{sec:gmn}

Generative matching networks aim to model conditional generative distributions of the form 
\begin{equation}\label{eq:generative_model}
    p(\obsvars | \Obsvars, \params) = \int p(\hidvars | \Obsvars, \params) p(\obsvars | \hidvars, \Obsvars, \params) d \hidvars,
\end{equation}
where $\hidvars$ is a latent variable generated by a (potentially data-dependent) prior $p(\hidvars | \Obsvars, \params)$ and $p(\obsvars | \hidvars, \Obsvars, \params)$ is a conditional likelihood.

We assume that the model is allowed to train on a large number of examples from a certain domain, accumulating knowledge about the \emph{domain} in parameters $\params$.
At the test time, the model can be conditioned on an additional dataset $\Obsvars = \{ \obsvars_1, \obsvars_2, \ldots, \obsvars_T \}$ and has to adapt it's generative distribution to the conditioning data.

In order to design a fast adaptation mechanism, we have to make certain assumptions about relationships between training data and the new data $\Obsvars$ used to condition the model. 
Thus we assume the homogeneity of generative processes for training and conditioning data up to some parametrization.
The generative process is assumed to have an approximately linear dependence on such parameters, i.e. the interpolation between parameters corresponding to different examples of the same concept can serve as good parameters for generating other examples.



\subsection{Basic model}\label{sec:basic_model}

\begin{figure*}[t]
\begin{center}
\includegraphics[width=\textwidth]{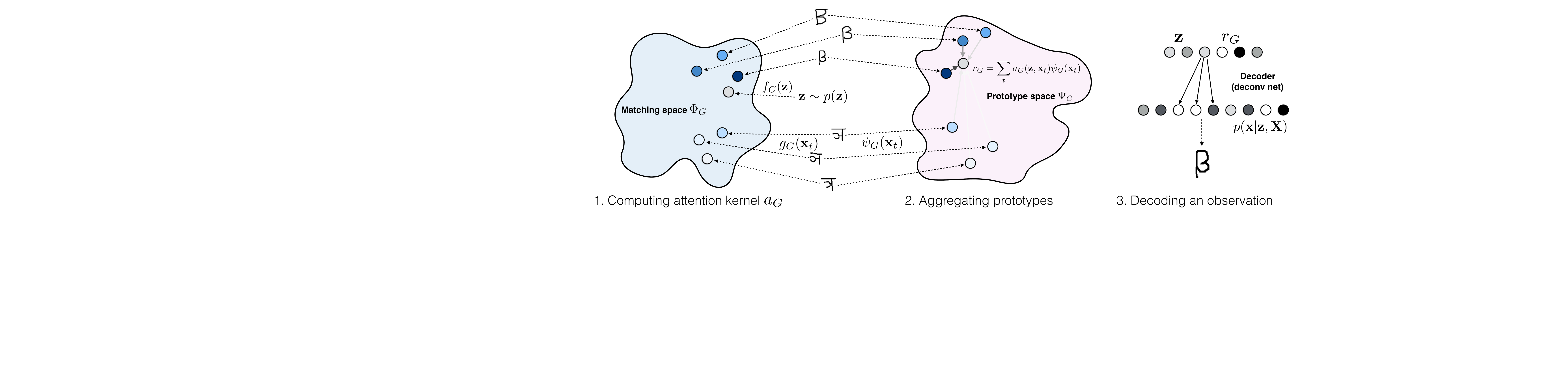}
\end{center}
\caption{Generation of a new sample in a basic generative matching network, see section~\ref{sec:basic_model} for the description of functions $f$, $g$ and $\psi$.}
\label{fig:simple_gmn}
\end{figure*}

In the basic version of our model, the prior is simply a standard Normal distribution $p(\hidvars) = \mathcal{N}(z | 0, I)$ which does not depend on the conditioning data $\Obsvars = \{ \obsvars_1, \ldots, \obsvars_T \}$.
A new observation is generated by first sampling the latent variable from the prior $\hidvars \sim p(\hidvars)$ and then matching it with each of the conditioning objects $\obsvars' \in \Obsvars$ to extract few relevant examples that would be used as prototypes for generation.

The relevance of the conditioning objects $\obsvars' \in \Obsvars$ is determined by a similarity function $\text{sim}(\cdot, \cdot)$.
Since latent variables and observations typically have very different representations, we first project them to the same \emph{matching space} $\Phi_G$, where similarity can be computed naturally.
A particularly convenient choice of such $\Phi_G$ would be a high-dimensional real vector space where the similarity function can be defined simply as cosine between corresponding vectors or, even simpler, dot-product which we used in our implementation.

The obtained similarity scores are then transformed through a softmax function defining the \emph{attention kernel} $a_G(\hidvars, \obsvars)$:
\begin{equation}\label{eq:simple_matching}
a_G(\hidvars, \obsvars_t) = \frac{\exp(\text{sim}(f_G(\hidvars), g_G(\obsvars_t)))}{\sum_{t'=1}^T \exp(\text{sim}(f_G(\hidvars), g_G(\obsvars_{t'})))}, \quad \result_G = \sum_{t=1}^T a_G(\hidvars, \obsvars_t) \psi_G(\obsvars_t),
\end{equation}
where functions $f_G(\cdot)$ and $g_G(\cdot)$ implemented by neural networks map observations and latent variables to the matching space $\Phi_G$.

The attention kernel provides a normalized weight assigned to a conditioning object which is used to extract different subsets of $\Obsvars$ conditioned on the sample $\hidvars$ in a soft manner.
These subsets are aggregated in $\result_G$ by interpolating between their \emph{prototype} descriptions computed with another network $\psi_G(\cdot)$.
Finally, the decoder network is provided with the weighted average $\result_G$ together with the latent variable $\hidvars$ and outputs a distribution over observations.
Figure \ref{fig:simple_gmn} contains a cartoon explanation of the described generative process.

Functions $g_G(\cdot)$ and $\phi_G(\cdot)$ are applied to observations and hence can be considered as feature extractors.
Since features useful to specify the generative process are not necessarily good for discrimination and vice versa, it makes sense to represent these functions differently, taking into account nature of a considered domain.
Thus, in our experiments these functions were implemented as convolutional networks with partial parameter sharing between them.

A major difference between the GMNs and the originally proposed discriminative matching networks~\citep{vinyals2016matching} is that since no label information is available to the model, the interpolation in equations~\eqref{eq:simple_matching} is performed not in the label space but rather in the prototype space $\Psi$ which itself is defined by the model and is learned during the training.

\subsection{Recognition model}

The recognition model $q(\hidvars | \Obsvars, \obsvars)$ is used to approximate the posterior distribution of latent variable given object $\obsvars$ in the context of previously observed conditioning objects $\Obsvars$. 
Similarly to the conditional likelihood, a similarity function operating in a (potentially different) matching space $\Psi_R$ is used to form an attention kernel, with the difference that function $f_R$ is now applied to the observation $\obsvars$.

The attention kernel and the interpolated prototype are given by:
\begin{equation}\label{eq:simple_recognition_matching}
a_R(\obsvars, \obsvars_t) = \frac{\exp(\text{sim}(f_R(\obsvars), g_R(\obsvars_t)))}{\sum_{t'=1}^T \exp(\text{sim}(f_R(\obsvars), g_R(\obsvars_{t'})))}, \quad r_R = \sum_{t=1}^T a_R(\obsvars, \obsvars_t) \psi_R(\obsvars_t)
\end{equation}
After the matching, interpolated prototype vector $r_R$ is used to compute parameters of the approximate posterior which in our case was a normal distribution with diagonal covariance matrix, i.e. $q(\hidvars | \Obsvars, \obsvars, \varparams) = \mathcal{N}(\hidvars | \mu(r_R), \Sigma(r_R))$.

\subsection{Pseudo-inputs}\label{sec:pseudo_inputs}

One can note that the described  model is not applicable in a situation where no conditioning objects are available, i.e. $\Obsvars=\emptyset$. 
A possible solution to this problem involves implicit addition of a \emph{pseudo-input} to the set of conditioning objects, i.e. $\Obsvars$.
There is no need to model pseudo-input as an actual observation, so we just represent it as corresponding outputs of functions $g^* = g^*(\obsvars^*)$ and $\psi^* = \psi^*(\obsvars^*)$ which are assumed to be another trainable parameters. 


\subsection{Full context matching}\label{sec:fce}

The potential limitation of the basic matching procedure~\eqref{eq:simple_matching} is that conditioning observations $\Obsvars$ are embedded independently from each other. 
Similarly to discriminative matching networks we address this problem by computing \emph{full context embeddings}~\citep{vinyals2015order}, i.e. embeddings that are computed jointly in the context of other conditioning examples.

We make $K$ attentional passes over $\Obsvars$ of the form~\eqref{eq:simple_matching}, guided by a recurrent controller $R$ which accumulates global knowledge about the conditioning data in its hidden state $\hstate$.
The hidden state is thus passed to feature extractors $f$ and $g$ in order to obtain context-dependent embeddings.

We refer to this process as the \emph{full context matching} procedure which modifies equation~\eqref{eq:simple_matching} as:
\begin{equation}\label{eq:fce_matching}
    a_G(\hidvars, \hstate_k, \obsvars_t) = \frac{\exp(\text{sim}(f_G(\hidvars, \hstate_k), g_G(\obsvars_t, \hstate_k)))}{\sum_{t'=1}^T \exp(\text{sim}(f_G(\hidvars, \hstate_k), g_G(\obsvars_{t'}, \hstate_k)))},
\end{equation}
where the interpolated prototype vector $\result_K$ and the hidden state $\hstate_k$ are given by:
\begin{align}\label{eq:fce_prototype}
    \textstyle \result_G^{k+1} = \sum_{t=1}^T a_G(\hidvars, \hstate_k, \obsvars_t) \psi_G(\obsvars_t, \hstate_k), \quad
    \hstate_{k+1} = R(\hstate_k, \result^k_G)
\end{align}
The output of the full matching procedure is thus the interpolated prototype vector from the last iteration $\result^K_G$ and the last hidden state of $\hstate_{K+1}$ passed to a decoder.
The analogous procedure is used in the recognition model.
In our implementation we shared the recurrent controller for generative and recognition models, thus we further refer to it as \emph{shared controller}.

\subsection{Data-dependent prior}

Full context matching described in the previous section also allows us to implement a context-dependent prior $p(\hidvars | \Obsvars, \params)$ which adjusts our prior assumptions based on the conditioning data $\Obsvars$.
Although in theory it should be possible to use a data-independent prior and translate the dependency  on the conditioning data to the likelihood with the same effect, as we show below in our experiments, data-dependent prior has a positive effect on model's performance.

We model the prior as a normal distribution with diagonal covariance whose parameters depend on the conditioning data $\Obsvars$ through the full context embedding~\eqref{eq:fce_matching}, \eqref{eq:fce_prototype}.
Since no query object such as a latent variable is available at the prior construction, only hidden state of the recurrent controller is used for matching.

\section{Training}\label{sec:training}

Training of our model consists of maximizing marginal likelihood of a dataset $\Obsvars$ which can be expressed as:
\begin{equation}\label{eq:marginal_likelihood}
    p(\Obsvars | \params) = \prod_{t=1}^T p(\obsvars_t | \Obsvars_{<t}, \params), \quad \Obsvars_{<t} = \{ \obsvars_s \}_{s=1}^{t-1}.
\end{equation}
We use the available training data to dynamically construct a large number of randomized few-shot learning problems and train GMNs to adapt on each of these problems simultaneously.
Such a training strategy is rooted in curriculum learning~\citep{bengio2009curriculum} and meta-learning~\citep{thrun1998lifelong, vilalta2002perspective, hochreiter2001learning}. 
It recently was successfully applied for one-shot discriminative learning~\citep{santoro2016one} and below we adapt it to our setting.

We define a \emph{task-generating} distribution $p_d(\Obsvars)$ which samples datasets $\Obsvars$ of size $T$ from training data.
Then we train our model to maximize the marginal likelihood of each dataset sampled on average:
$
\mathbb{E}_{p_d(\Obsvars)} \left[ \log p(\Obsvars | \params) \right] \rightarrow \max_{\params}.
$

A standard practice is constrain $p_d$ to generate datasets that consist only of objects of a single class, so that the model has a clear incentive to re-use conditioning data~\citep{rezende2016one, edwards2016towards}.
As we show further, Generative Matching Networks impose much less requirements for the data generating distribution and can be trained on datasets representing $C > 1$ different concepts or classes. 

Since the marginal likelihood~\eqref{eq:marginal_likelihood} as well as the conditional marginal likelihoods are intractable we instead use variational lower bound (see section~\ref{sec:var_inf}) as a proxy to $\log p(\Obsvars | \params)$:
\begin{equation*}
\textstyle \mathcal{L}(\Obsvars, \params, \varparams) = \sum_{t=1}^T \mathbb{E}_{q(\hidvars_t | \obsvars_t, \Obsvars_{<t}, \varparams)} [ \log p(\obsvars_t, \hidvars_t | \Obsvars_{<t}, \params) - \log q(\hidvars_t | \obsvars_t, \Obsvars_{<t}, \varparams) ].
\end{equation*}

\section{Experiments}\label{sec:experiments}

For our experiments we use the Omniglot dataset~\citep{lake2015human} which consists of 1623 classes of handwritten characters from 50 different alphabets.
The first 30 alphabets are devoted for training and the remaining 20 alphabets are left for testing.
Importantly, only 20 examples of each class are available which makes this dataset specifically useful for few-shot learning problems.
Unfortunately, the literature is inconsistent in usage of the dataset and multiple versions of Omniglot were used for evaluation which differ by train/test split, resolution, binarization and augmentation, see e.g.~\citep{burda2015importance, rezende2016one, santoro2016one}.

We use the canonical split provided by~\citet{lake2015human}. 
In order to speed-up training we downscaled images to $28 \times 28$ resolution and since the result was fully binary we did not apply any further pre-processing. 
We also did not augment our data in contrast to~\citep{santoro2016one, edwards2016towards} to make future comparisons with our results easier. Unless otherwise stated, we train models on datasets of length $T=20$ and of up to $C_{\text{train}}=2$ different classes as we did not observe any improvement from training on more diverse datasets.

\subsection{Fast learning and few-shot generation}\label{sec:few_shot}

In this section we compare generative matching networks with a set of baselines by expected conditional likelihoods $\mathbb{E}_{p_d(\Obsvars)} p(\obsvars_t | \Obsvars_{<t})$.
The conditional likelihoods were estimated using importance sampling with $1000$ samples from the recognition model used as a proposal.


\paragraph{Models} We compare different variants of generative matching networks with a set of baselines.
To make the evaluation consistent, all the models use the same architecture for the encoder and the decoder parts, which we describe in detail in the supplementary material.

\textbf{VAE.} In order to get a sense of quantitative improvement over non-conditional generative models, we implemented a variational auto-encoder (VAE) with a similar architecture to our model, but lacking any adaptation mechanisms.\\
\textbf{One-shot VAE.} Following~\citet{rezende2016one}, we implemented a simple conditional VAE of the form $p(\obsvars | \obsvars') = \int p(\hidvars | \obsvars') p(\obsvars | \hidvars, \obsvars') d\hidvars$.
This model aims at generating a new example of a character represented by $\obsvars'$ and hence can be trained only with $C=1$. \\
\textbf{Neural Statistician.} Another non-trivial baseline is the Neural Statistician model~\citep{edwards2016towards} (see section ~\ref{sec:adaptation}) which can be considered as state of the art in few-shot generative modelling.
\\
\textbf{Generative Matching Network (GMN).} For evaluation, we used a model with full matching procedure (see section~\ref{sec:fce}) using $4$ steps for the shared controller and a single step for the prior controller (see supplementary materials for discussion of these parameters). 
Further, if not stated otherwise, we consider GMNs with a single pseudo-input (described in section ~\ref{sec:pseudo_inputs}), as we found the benefit of adding more pseudo-inputs negligible.\\
\textbf{Generative Matching Network without attention (GMN, no attention).} We also consider a restricted version of our model with the attentional matching procedure~\eqref{eq:simple_matching} replaced by a uniform kernel that effectively leads to simple averaging prototypes of all conditioning examples.
The same architecture was used in the neural statistician model to aggregate conditioning data, except that the latter models uncertainty about this aggregate.


\begin{table*}[t]
\setlength{\tabcolsep}{5pt}
\caption{Conditional negative log-likelihoods for the test part of Omniglot. $C_{\text{train}}$ and $C_{\text{test}}$ denote the maximum number of classes in datasets used for training and evaluating respectively.}
\label{tbl:predictive}
\centering
\setlength{\tabcolsep}{.35em}
\begin{tabular}{lccccccccc}
\textbf{}                             & \multicolumn{1}{l}{\textbf{}} & \multicolumn{8}{c}{\textsc{\textbf{Number of conditioning examples}}}                                                            \\
\textsc{\textbf{Model}}                        & \textbf{$C_{\text{test}}$}               & \textsc{0} & \textsc{1} & \textsc{2} & \textsc{3} & \textsc{4} & \textsc{5} & \textsc{10} & \textsc{19} \\
 \hline  
GMN, $C_{\text{train}}=2$            & 1                             & \bf 89.7       & \bf 83.3       & \bf 78.9       & \bf 75.7       & \bf 72.9       & \bf 70.1       & \bf 59.9        & \bf 45.8        \\
GMN, $C_{\text{train}}=2$            & 2                             & \bf 89.4       & \bf 86.4       & \bf 84.9       & \bf 82.4       & \bf 81.0       & \bf 78.8       & \bf 71.4        & \bf 61.2        \\
GMN, $C_{\text{train}}=2$            & 3                             & \bf 89.6       & \bf 88.1       & \bf 86.0       & \bf 85.0       & \bf 84.1       & \bf 82.0       & \bf 76.3        & \bf 69.4        \\
GMN, $C_{\text{train}}=2$            & 4                             & \bf 89.3       & \bf 88.3       & \bf 87.3       & \bf 86.7       & \bf 85.4       & \bf 84.0       & \bf 80.2        & \bf 73.7        \\
GMN, $C_{\text{train}}=2$, no pseudo-input      & 1                             &            & 93.5       & 82.2       & 78.6       & 76.8       & 75.0       & 69.7        & 64.3        \\
GMN, $C_{\text{train}}=2$, no pseudo-input      & 2                             &            &            & 86.1       & 83.7       & 82.8       & 81.0       & 76.5        & 71.4        \\
GMN, $C_{\text{train}}=2$, no pseudo-input      & 3                             &            &            &           & 86.1       & 84.7       & 83.8       & 79.7        & 75.3        \\
GMN, $C_{\text{train}}=2$, no pseudo-input      & 4                             &            &            &            &            & 86.8       & 85.7       & 82.5        & 78.0        \\
\hline
VAE                                  &                               & 89.1       &            &            &            &            &            &             &             \\
One-shot VAE & 1                              &        & 83.9           &            &            &            &            &             &             \\
Neural statistician, $C_{\text{train}}=1$       & 1                             &        & 102       & 83.4       & 77.8       & 75.2       & 74.6       & 71.7        & 71.5        \\
Neural statistician, $C_{\text{train}}=2$       & 2                             &        &        & 86.4       & \textbf{82.2}      & 82.3       & 80.6       &  79.7       & 79.0        \\
GMN, $C_{\text{train}}=1$, no attention       & 1                             & 92.4       & 84.5       & 82.3       & 81.4       & 81.1       & 80.4       & 79.8        & 79.7        \\
GMN, $C_{\text{train}}=2$, no attention       & 2                             & 88.2       & 86.6       & 86.4       & 85.7       & 85.3       & 84.5       & 83.7        & 83.4        \\
GMN, $C_{\text{train}}=1$, no attention, no pseudo-input & 1                             &            & 88.0       & 84.1       & 82.9       & 82.4       & 81.7       & 80.9        & 80.7        \\
GMN, $C_{\text{train}}=2$, no attention, no pseudo-input & 2                             &            &            & 85.7       & 85.0       & 85.3       & 84.6       & 84.5        & 83.7      
\end{tabular}
\end{table*}

Table~\ref{tbl:predictive} contains results of the evaluation on test alphabets from Omniglot.
One can see, that Generative Matching Networks demonstrate significant improvement in predictive performance as more data is available to the model.
As one would expect, larger values of $C_{\text{test}}$ made adaptation harder since on average less examples of the same class are available to the model. 
Still GMNs are capable of working in low-data regime even when $C_{\text{test}} > C_{\text{train}}$.

Unsurprisingly, models with prototype averaging (GMN, no attention and Neural statistician) performed well for simple datasets constructed of a single class, although significantly worse than the proposed matching procedure. 
On more difficult datasets with mixed examples of two different classes ($C_{\text{test}} = 2$) GMNs clearly outperformed all concurrent models, thus justifying importance of nonparametric representations for complex data.

\begin{figure}
    \begin{tabular}{cc}
\includegraphics[width=2.5in, height=1.38in]{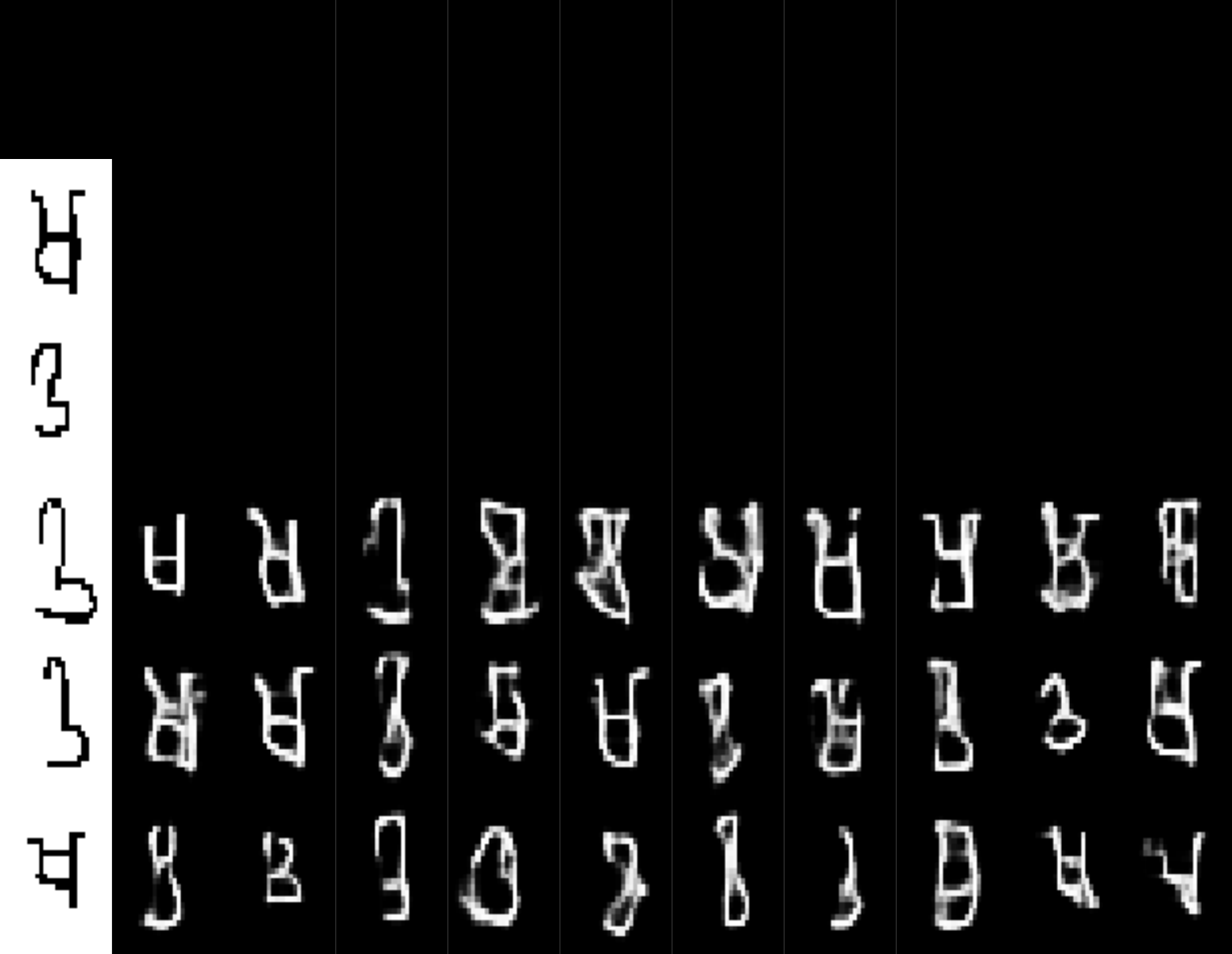} & \includegraphics[width=2.5in]{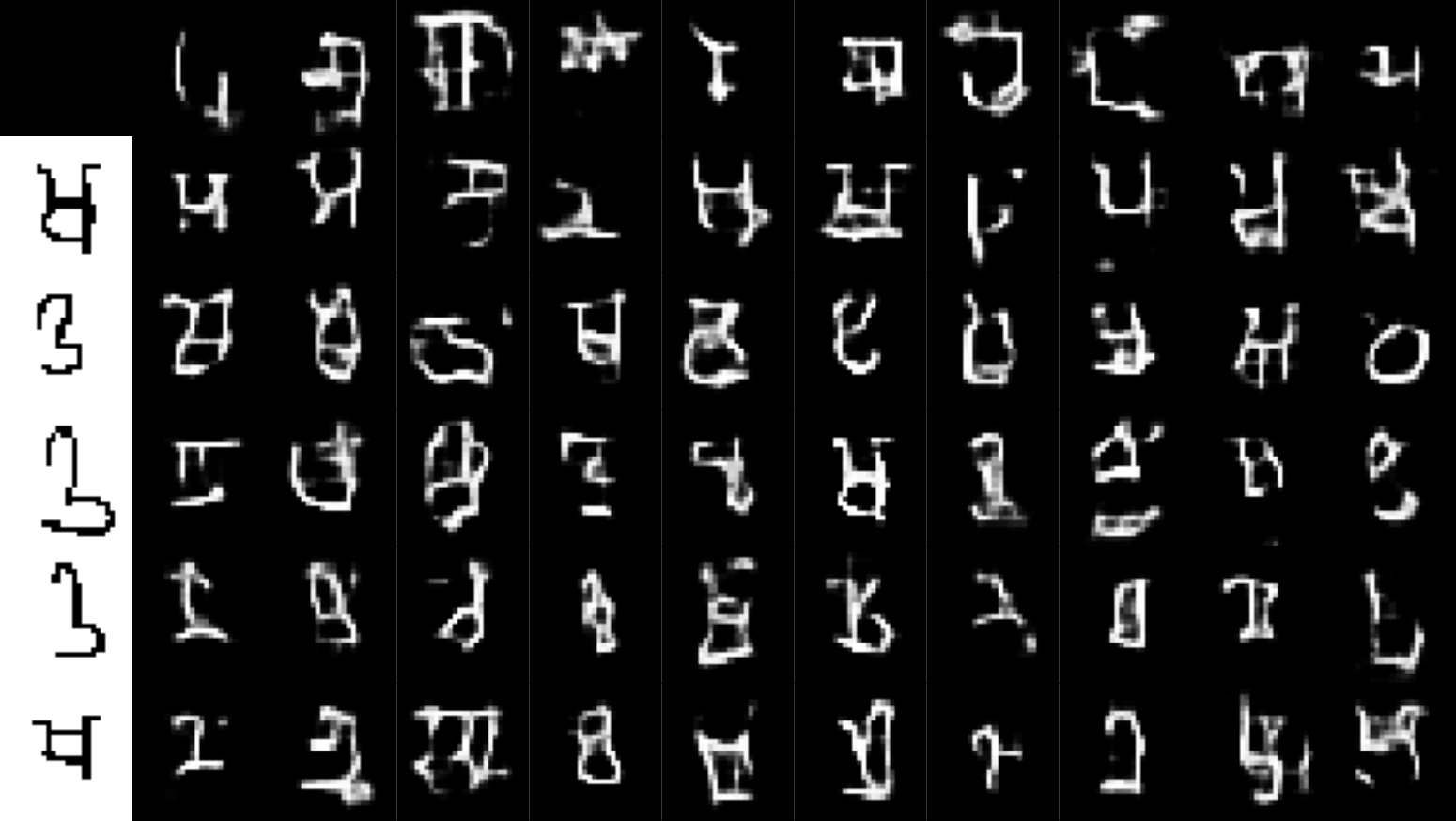} \\
\includegraphics[width=2.5in]{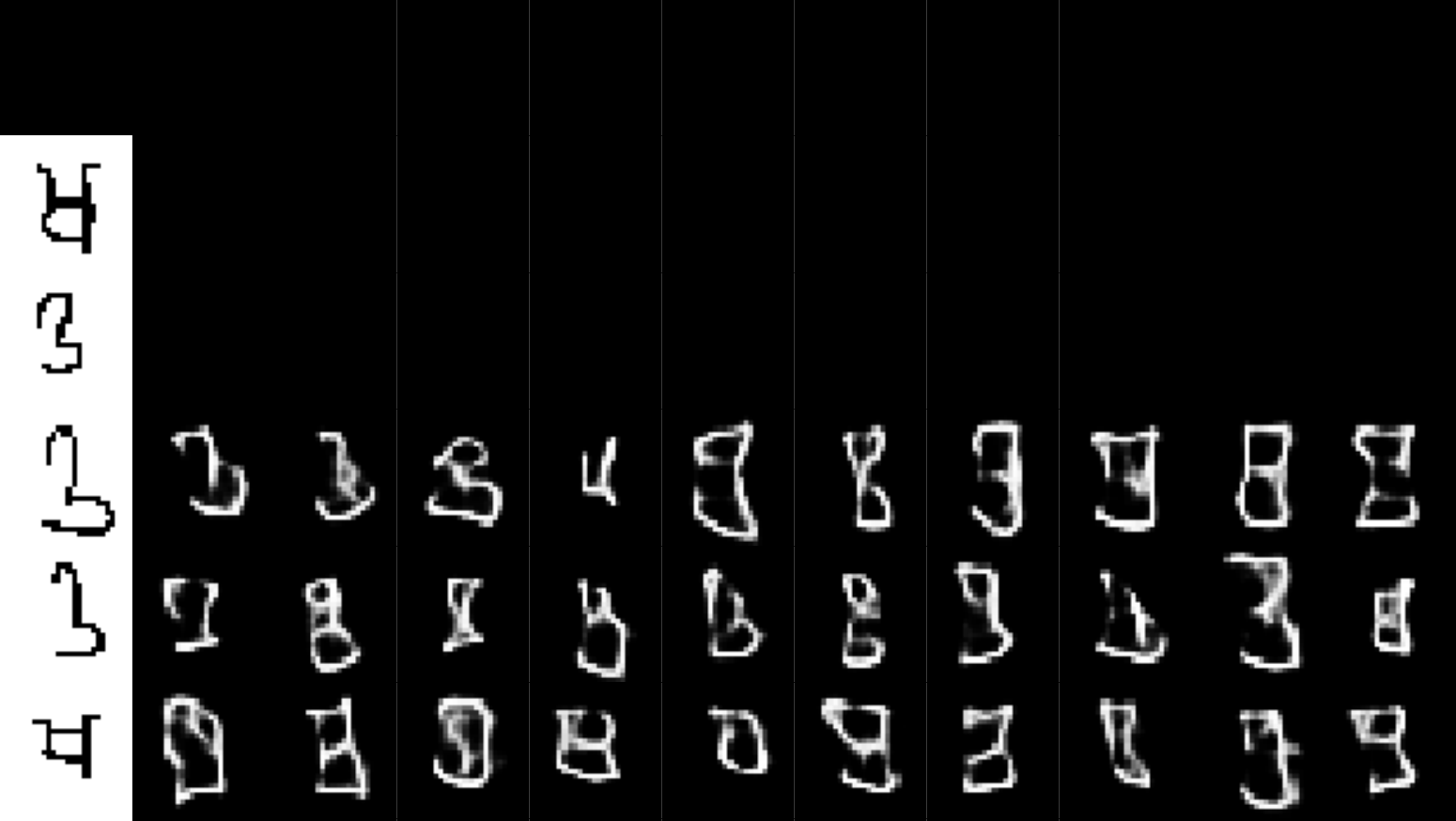} & \includegraphics[width=2.5in]{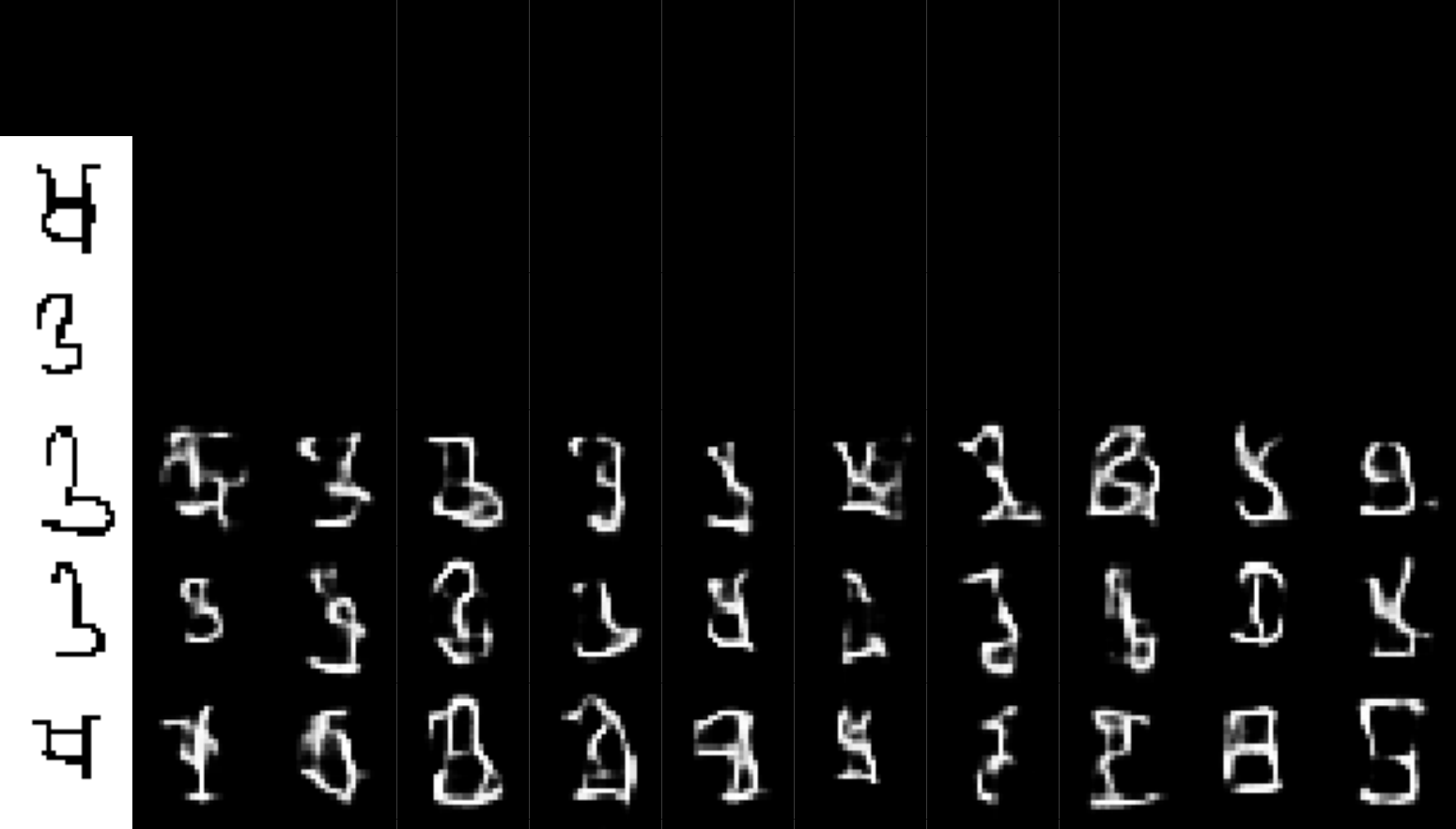} \\
\end{tabular}
    \caption{Conditionally generated samples. Left-right, top-bottom: GMN (no pseudo-input), GMN (one pseudo-input),  GMN (no attention, no pseudo-input), Neural statistician.
    For each image: first column contains conditioning data in the order it is revealed to the model. Row number $t$ (counting from zero) consists of samples conditioned on first $t$ input examples.}\label{fig:samples}
\end{figure}

In order to visually assess the fast learning ability of GMNs we also provide conditionally generated samples in figure~\ref{fig:samples}. 
Interestingly, the GMN without pseudo-inputs generated samples more similar to the conditioning data while sacrificing the predictive performance.
Such counter-intuitive mismatch between visual quality of samples and predictive performance has been studied before~\citep{theis2015note} and may suggest that without a pseudo-input, GMNs tend to learn less of the common knowledge about the domain and slightly ``overfit'' to conditioning data.
Therefore, presence or absence of the pseudo-input should depend on the target application of interest, i.e. density estimation or producing new examples.
We provide more samples generated by GMNs in supplementary materials.

\subsection{Classification}

Even if an adaptive generative model did not take into account the label information during the training, as in the case of GMN, it can still be used for few-shot classification.
Given a small number of labeled examples $\Obsvars_c = \{ \obsvars_{c, 1}, \obsvars_{c, 2}, \ldots \obsvars_{c, N} \}$ for each class $c \in \{ 1, \ldots, C \}$, it possible to use $p(\obsvars | \Obsvars_c)$ as a score for assigning label $c$ to $\obsvars$.
While one would not expect this method to provide state of the art results, few-shot classification using class-conditional densities is still a good test for adaptation capabilities.

Alternatively, one may use the recognition model $q(\hidvars | \obsvars, \Obsvars_1, \ldots, \Obsvars_{C})$ to extract features describing the new object $\obsvars$ and then use a classifier of choice, e.g. nearest-neighbour classifier with cosine similarity of mean parameters as in our experiments.
This method can be used as a general adaptive feature extraction technique which itself is an interesting application of GMNs.
In the neural statistician model $q(\globvars | \obsvars)$ and $ q(\globvars | \Obsvars_c)$ were used for feature extraction.

The results under different number of training examples available are provided in table~\ref{tbl:classification}. 
Surprisingly, the simpler GMN with uniform attention outperformed all other models, including the full model. 
As shown in the previous section, GMNs without pseudo-inputs are very smooth density estimators, so, perhaps, even conditioned on a number of same-class examples, they still assign enough probability mass to discrepant observations.

\begin{table}[ht]
\centering
\caption{Few-shot classification accuracy (\%) on the test part of Omniglot. All models were trained with $C_{\text{train}}=1$ and no pseudo-input (if applicable). }
\label{tbl:classification}
\begin{tabular}{llcccc}
\textbf{}                   & \multicolumn{1}{l}{\textbf{}}       & \multicolumn{2}{c}{\textsc{\textbf{5 classes}}}                                        & \multicolumn{2}{c}{\textsc{\textbf{20 classes}}}                                       \\
\textsc{\textbf{Model}}              & \multicolumn{1}{l}{\textsc{\textbf{Method}}} & \multicolumn{1}{l}{\textsc{\textbf{1-shot}}} & \multicolumn{1}{l}{\textsc{\textbf{5-shot}}} & \multicolumn{1}{l}{\textsc{\textbf{1-shot}}} & \multicolumn{1}{l}{\textsc{\textbf{5-shot}}} \\
\hline 
GMN  &  likelihood                          & 82.7                                & \textbf{97.4}                                & 64.3                                & \textbf{90.8}                                \\
GMN, no attention    & likelihood                          & \textbf{90.8}                                & 96.7                                & \textbf{77.0}                                & \textbf{91.0}                                \\
GMN & cosine                         & 62.7                                & 80.8                                & 45.1                                & 67.2                                \\
GMN, no attention     & cosine                         & 72.0                                & 86.0                                & 50.1                                & 72.6                                \\
One-shot VAE & likelihood & \textbf{90.2} & & 76.3 &  \\
One-shot VAE & cosine & 72.1 & & 50.1 & \\
Neural statistician & likelihood & 82.0 & 94.8 & 63.1 & 87.6 \\
Neural statistician & cosine & 66.4 & 85.5 & 47.3 & 71.7 \\
\hline
Matching networks, no fine-tuning & cosine & 98.1 & 98.9 & 93.8 & 98.5 
\end{tabular}
\end{table}

\section{Conclusion}\label{sec:conclusion}

We presented Generative Matching Network, a new conditional generative model that is capable of fast adaptation to conditioning dataset by adjusting both the latent space and the predictive density.
The nonparametric matching procedure enabling these features can be seen as a generalization of the original matching network architecture since it allows the model to define the label space itself, thus extending applicability of matching networks to unsupervised and perhaps semi-supervised settings.
We believe that these ideas can evolve further and help to implement more data-efficient models in other domains such as reinforcement learning where data acquisition is especially hard. 

\subsubsection*{Acknowledgments}

We would like to thank Michael Figurnov and Timothy Lillicrap for useful discussions.
Dmitry P. Vetrov is supported by RFBR project No.15-31-20596 (mol-a-ved) and by Microsoft: MSU joint research center (RPD 1053945).

\bibliography{example_paper}

\begin{thebibliography}{26}
\providecommand{\natexlab}[1]{#1}
\providecommand{\url}[1]{\texttt{#1}}
\expandafter\ifx\csname urlstyle\endcsname\relax
  \providecommand{\doi}[1]{doi: #1}\else
  \providecommand{\doi}{doi: \begingroup \urlstyle{rm}\Url}\fi

\bibitem[Bengio et~al.(2009)Bengio, Louradour, Collobert, and
  Weston]{bengio2009curriculum}
Bengio, Yoshua, Louradour, J{\'e}r{\^o}me, Collobert, Ronan, and Weston, Jason.
\newblock Curriculum learning.
\newblock In \emph{Proceedings of the 26th annual international conference on
  machine learning}, pp.\  41--48. ACM, 2009.

\bibitem[Burda et~al.(2015)Burda, Grosse, and
  Salakhutdinov]{burda2015importance}
Burda, Yuri, Grosse, Roger, and Salakhutdinov, Ruslan.
\newblock Importance weighted autoencoders.
\newblock \emph{arXiv preprint arXiv:1509.00519}, 2015.

\bibitem[Chung et~al.(2015)Chung, G{\"{u}}l{\c c}ehre, Cho, and
  Bengio]{chung-icml15-gated}
Chung, Junyoung, G{\"{u}}l{\c c}ehre, {\c C}ağlar, Cho, Kyunghyun, and Bengio,
  Yoshua.
\newblock Gated feedback recurrent neural networks.
\newblock In \emph{Proceedings of the 32nd International Conference on Machine
  Learning (ICML'15)}, 2015.

\bibitem[Diaconis \& Freedman(1980)Diaconis and Freedman]{diaconis1980finite}
Diaconis, Persi and Freedman, David.
\newblock Finite exchangeable sequences.
\newblock \emph{The Annals of Probability}, pp.\  745--764, 1980.

\bibitem[Edwards \& Storkey(2016)Edwards and Storkey]{edwards2016towards}
Edwards, Harrison and Storkey, Amos.
\newblock Towards a neural statistician.
\newblock \emph{arXiv preprint arXiv:1606.02185}, 2016.

\bibitem[Gershman \& Goodman(2014)Gershman and Goodman]{gershman2014amortized}
Gershman, Samuel~J and Goodman, Noah~D.
\newblock Amortized inference in probabilistic reasoning.
\newblock In \emph{Proceedings of the 36th Annual Conference of the Cognitive
  Science Society}, 2014.

\bibitem[Goodfellow et~al.(2014)Goodfellow, Pouget-Abadie, Mirza, Xu,
  Warde-Farley, Ozair, Courville, and Bengio]{goodfellow2014generative}
Goodfellow, Ian, Pouget-Abadie, Jean, Mirza, Mehdi, Xu, Bing, Warde-Farley,
  David, Ozair, Sherjil, Courville, Aaron, and Bengio, Yoshua.
\newblock Generative adversarial nets.
\newblock In \emph{Advances in Neural Information Processing Systems}, pp.\
  2672--2680, 2014.

\bibitem[Gutmann \& Hyv{\"a}rinen(2012)Gutmann and
  Hyv{\"a}rinen]{gutmann2012noise}
Gutmann, Michael~U and Hyv{\"a}rinen, Aapo.
\newblock Noise-contrastive estimation of unnormalized statistical models, with
  applications to natural image statistics.
\newblock \emph{Journal of Machine Learning Research}, 13\penalty0
  (Feb):\penalty0 307--361, 2012.

\bibitem[He et~al.(2015)He, Zhang, Ren, and Sun]{he2015delving}
He, Kaiming, Zhang, Xiangyu, Ren, Shaoqing, and Sun, Jian.
\newblock Delving deep into rectifiers: Surpassing human-level performance on
  imagenet classification.
\newblock In \emph{Proceedings of the IEEE International Conference on Computer
  Vision}, pp.\  1026--1034, 2015.

\bibitem[Hochreiter et~al.(2001)Hochreiter, Younger, and
  Conwell]{hochreiter2001learning}
Hochreiter, Sepp, Younger, A~Steven, and Conwell, Peter~R.
\newblock Learning to learn using gradient descent.
\newblock In \emph{International Conference on Artificial Neural Networks},
  pp.\  87--94. Springer, 2001.

\bibitem[Jaakkola \& Jordan(2000)Jaakkola and Jordan]{jaakkola2000bayesian}
Jaakkola, Tommi~S and Jordan, Michael~I.
\newblock Bayesian parameter estimation via variational methods.
\newblock \emph{Statistics and Computing}, 10\penalty0 (1):\penalty0 25--37,
  2000.

\bibitem[Jordan et~al.(1999)Jordan, Ghahramani, Jaakkola, and
  Saul]{jordan1999introduction}
Jordan, Michael~I, Ghahramani, Zoubin, Jaakkola, Tommi~S, and Saul, Lawrence~K.
\newblock An introduction to variational methods for graphical models.
\newblock \emph{Machine learning}, 37\penalty0 (2):\penalty0 183--233, 1999.

\bibitem[Kingma \& Welling(2013)Kingma and Welling]{kingma2013auto}
Kingma, Diederik~P and Welling, Max.
\newblock Auto-encoding variational bayes.
\newblock \emph{arXiv preprint arXiv:1312.6114}, 2013.

\bibitem[Lake et~al.(2015)Lake, Salakhutdinov, and Tenenbaum]{lake2015human}
Lake, Brenden~M, Salakhutdinov, Ruslan, and Tenenbaum, Joshua~B.
\newblock Human-level concept learning through probabilistic program induction.
\newblock \emph{Science}, 350\penalty0 (6266):\penalty0 1332--1338, 2015.

\bibitem[McCloskey \& Cohen(1989)McCloskey and
  Cohen]{mccloskey1989catastrophic}
McCloskey, Michael and Cohen, Neal~J.
\newblock Catastrophic interference in connectionist networks: The sequential
  learning problem.
\newblock \emph{Psychology of learning and motivation}, 24:\penalty0 109--165,
  1989.

\bibitem[Ratcliff(1990)]{ratcliff1990connectionist}
Ratcliff, Roger.
\newblock Connectionist models of recognition memory: constraints imposed by
  learning and forgetting functions.
\newblock \emph{Psychological review}, 97\penalty0 (2):\penalty0 285, 1990.

\bibitem[Rezende et~al.(2014)Rezende, Mohamed, and
  Wierstra]{rezende2014stochastic}
Rezende, Danilo~J, Mohamed, Shakir, and Wierstra, Daan.
\newblock Stochastic backpropagation and approximate inference in deep
  generative models.
\newblock In \emph{Proceedings of the 31st International Conference on Machine
  Learning (ICML-14)}, pp.\  1278--1286, 2014.

\bibitem[Rezende et~al.(2016)Rezende, Mohamed, Danihelka, Gregor, and
  Wierstra]{rezende2016one}
Rezende, Danilo~Jimenez, Mohamed, Shakir, Danihelka, Ivo, Gregor, Karol, and
  Wierstra, Daan.
\newblock One-shot generalization in deep generative models.
\newblock \emph{arXiv preprint arXiv:1603.05106}, 2016.

\bibitem[Salakhutdinov et~al.(2013)Salakhutdinov, Tenenbaum, and
  Torralba]{salakhutdinov2013learning}
Salakhutdinov, Ruslan, Tenenbaum, Joshua~B, and Torralba, Antonio.
\newblock Learning with hierarchical-deep models.
\newblock \emph{IEEE transactions on pattern analysis and machine
  intelligence}, 35\penalty0 (8):\penalty0 1958--1971, 2013.

\bibitem[Santoro et~al.(2016)Santoro, Bartunov, Botvinick, Wierstra, and
  Lillicrap]{santoro2016one}
Santoro, Adam, Bartunov, Sergey, Botvinick, Matthew, Wierstra, Daan, and
  Lillicrap, Timothy.
\newblock One-shot learning with memory-augmented neural networks.
\newblock \emph{arXiv preprint arXiv:1605.06065}, 2016.

\bibitem[Snelson \& Ghahramani(2005)Snelson and Ghahramani]{snelson2005compact}
Snelson, Edward and Ghahramani, Zoubin.
\newblock Compact approximations to bayesian predictive distributions.
\newblock In \emph{Proceedings of the 22nd international conference on Machine
  learning}, pp.\  840--847. ACM, 2005.

\bibitem[Theis et~al.(2015)Theis, Oord, and Bethge]{theis2015note}
Theis, Lucas, Oord, A{\"a}ron van~den, and Bethge, Matthias.
\newblock A note on the evaluation of generative models.
\newblock \emph{arXiv preprint arXiv:1511.01844}, 2015.

\bibitem[Thrun(1998)]{thrun1998lifelong}
Thrun, Sebastian.
\newblock Lifelong learning algorithms.
\newblock In \emph{Learning to learn}, pp.\  181--209. Springer, 1998.

\bibitem[Vilalta \& Drissi(2002)Vilalta and Drissi]{vilalta2002perspective}
Vilalta, Ricardo and Drissi, Youssef.
\newblock A perspective view and survey of meta-learning.
\newblock \emph{Artificial Intelligence Review}, 18\penalty0 (2):\penalty0
  77--95, 2002.

\bibitem[Vinyals et~al.(2015)Vinyals, Bengio, and Kudlur]{vinyals2015order}
Vinyals, Oriol, Bengio, Samy, and Kudlur, Manjunath.
\newblock Order matters: Sequence to sequence for sets.
\newblock \emph{arXiv preprint arXiv:1511.06391}, 2015.

\bibitem[Vinyals et~al.(2016)Vinyals, Blundell, Lillicrap, Kavukcuoglu, and
  Wierstra]{vinyals2016matching}
Vinyals, Oriol, Blundell, Charles, Lillicrap, Timothy, Kavukcuoglu, Koray, and
  Wierstra, Daan.
\newblock Matching networks for one shot learning.
\newblock \emph{arXiv preprint arXiv:1606.04080}, 2016.

\end{thebibliography}
\bibliographystyle{icml2017}

\newpage

\section{More samples from the model}

\begin{figure*}
\begin{tabular}{cc}
\includegraphics[width=2.5in]{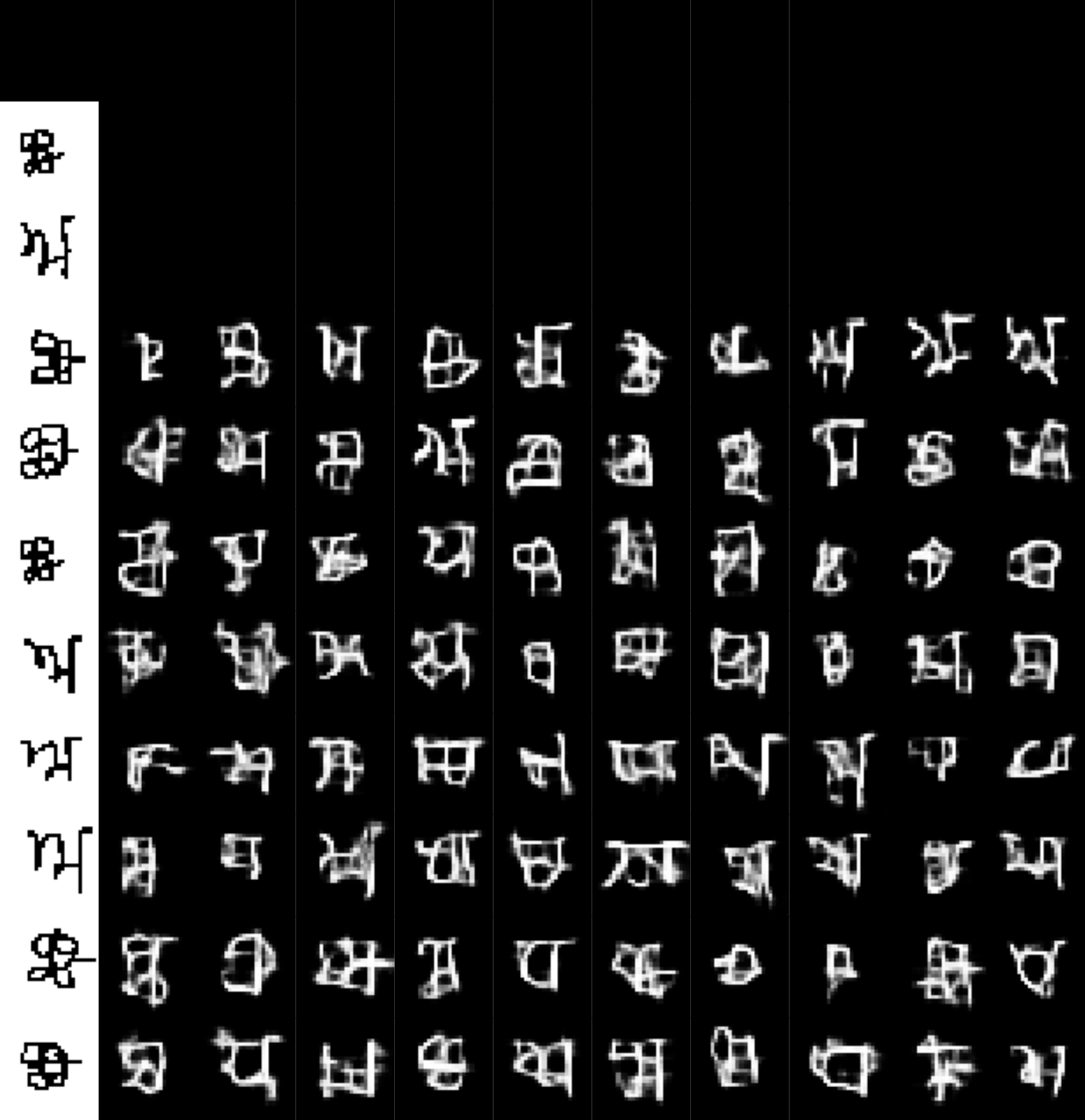} & \includegraphics[width=2.5in]{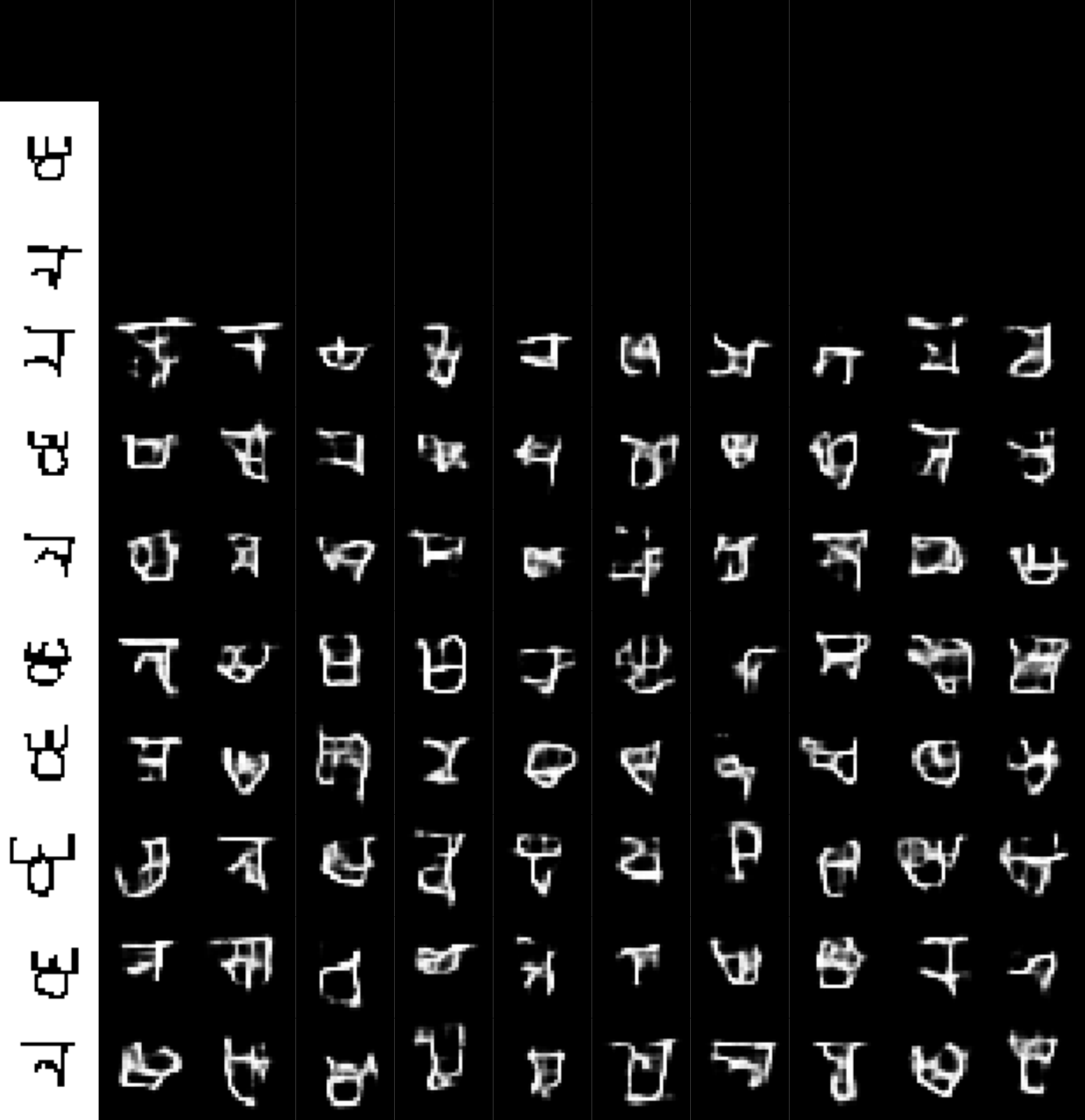} \\
\includegraphics[width=2.5in]{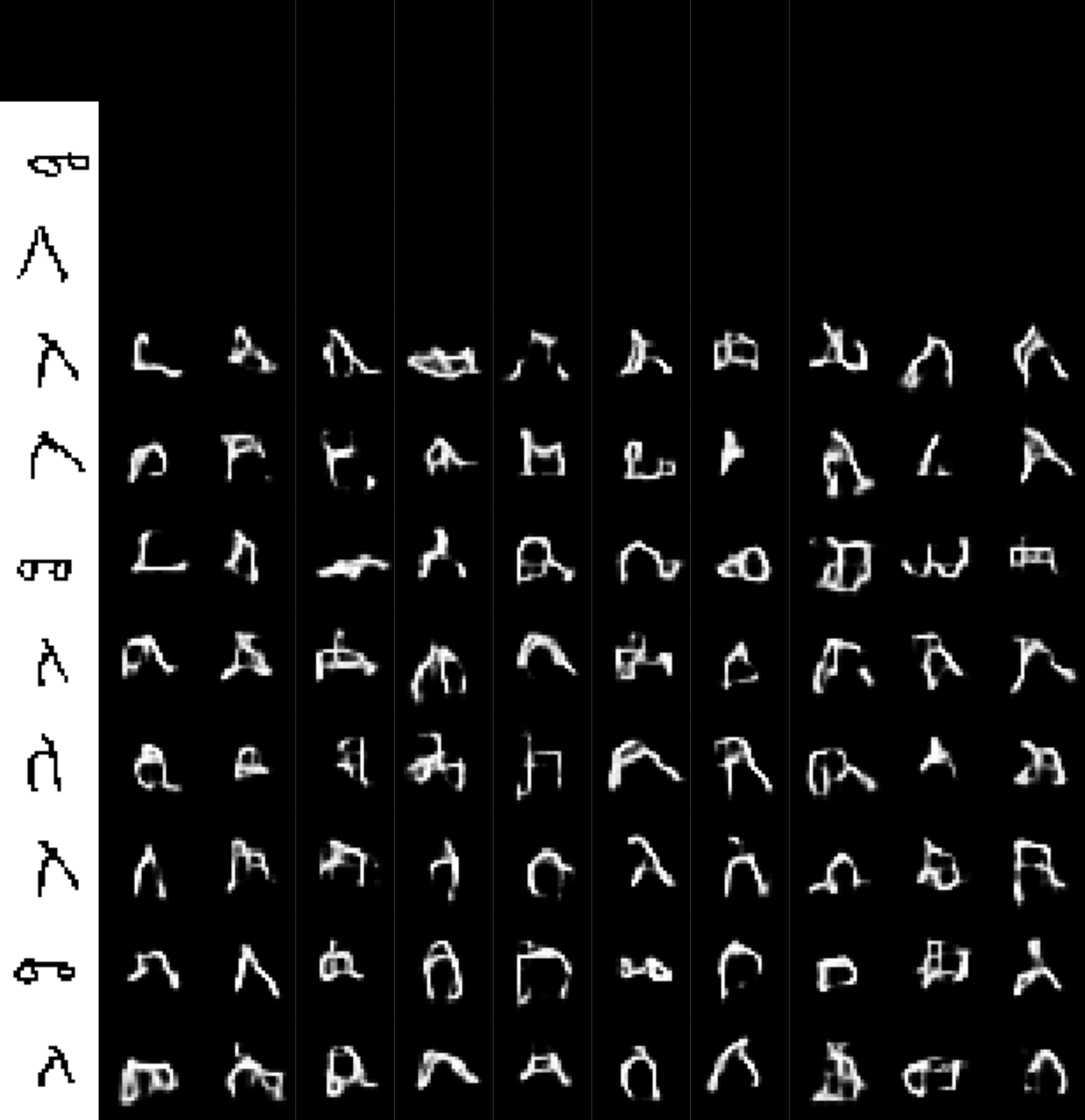} & \includegraphics[width=2.5in]{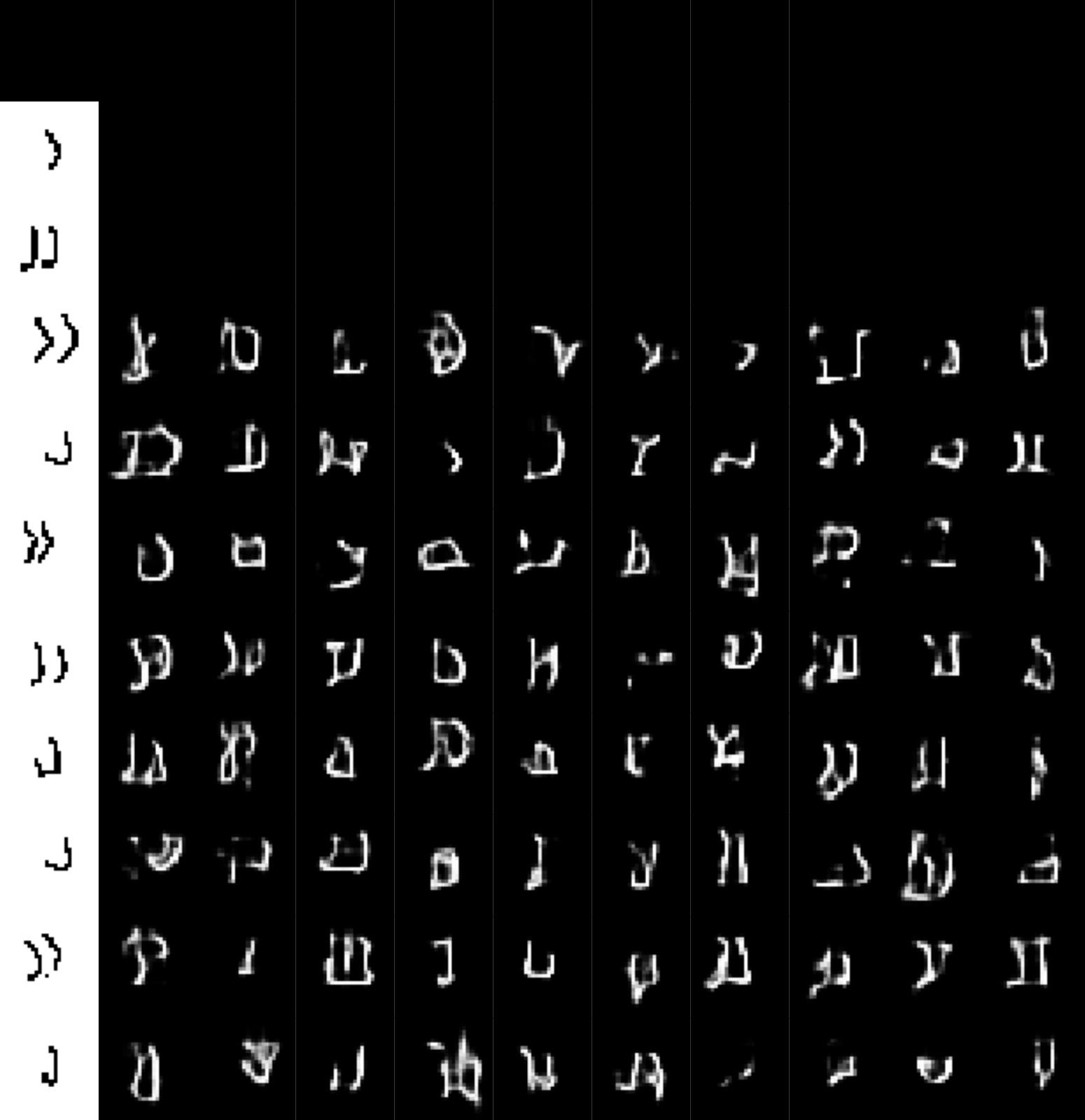} \\
\end{tabular}
\caption{Additional conditionally-generated samples from the GMN, no pseudo-inputs used.}
\label{fig:additional_samples_gmn_conditional}
\end{figure*}

\begin{figure*}
\begin{tabular}{cc}
\includegraphics[width=2.5in]{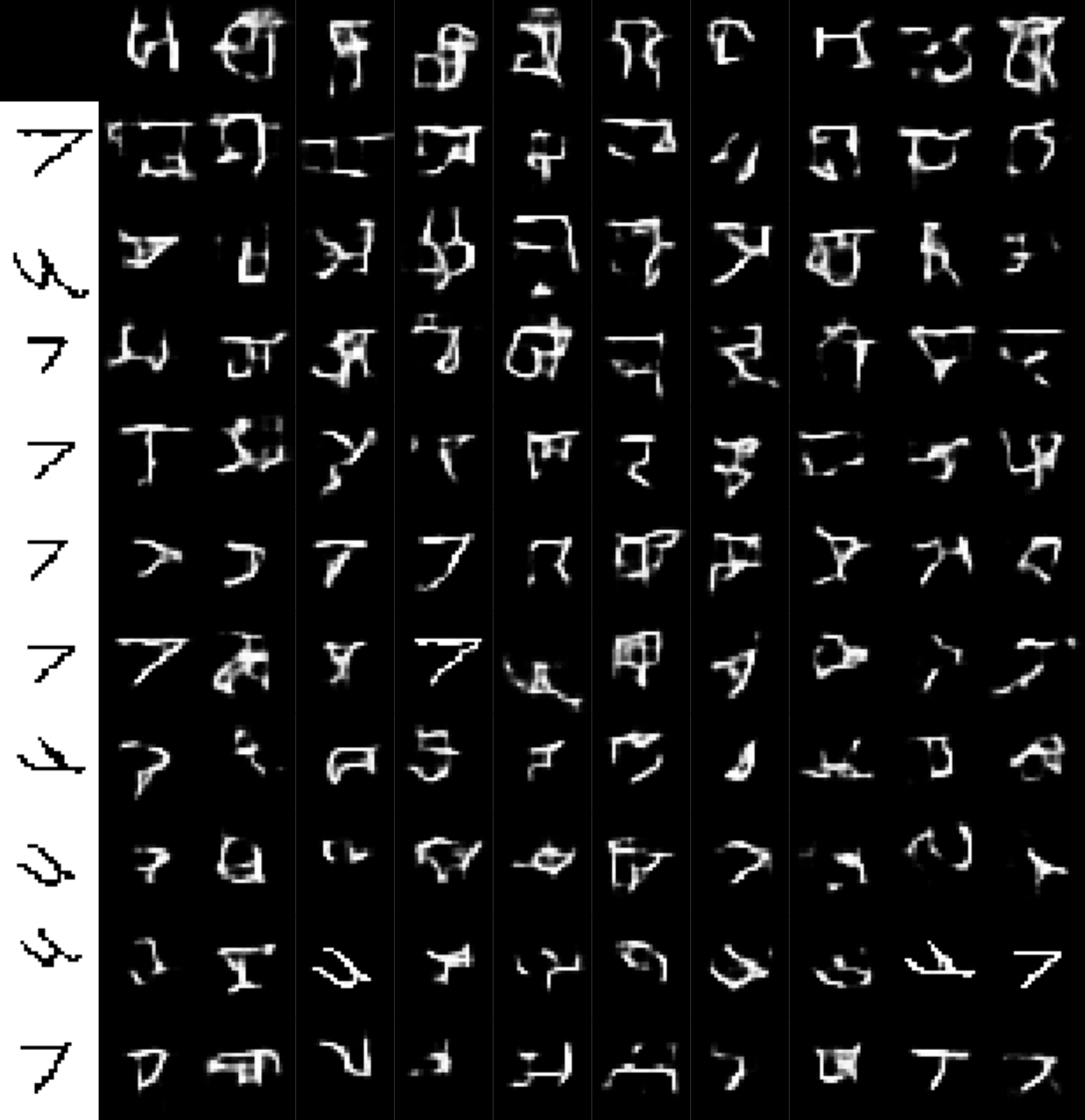} & \includegraphics[width=2.5in]{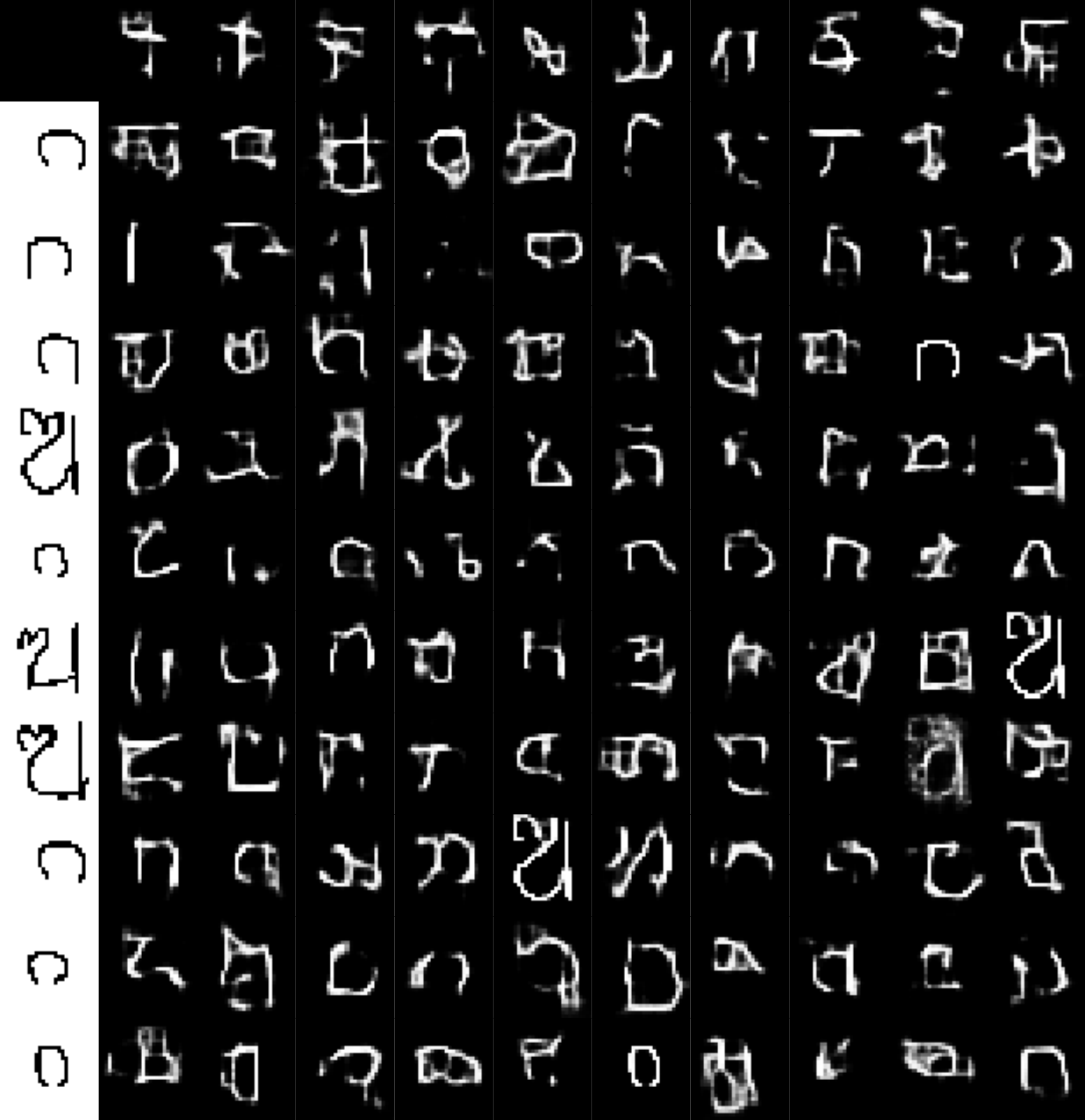} \\
\includegraphics[width=2.5in]{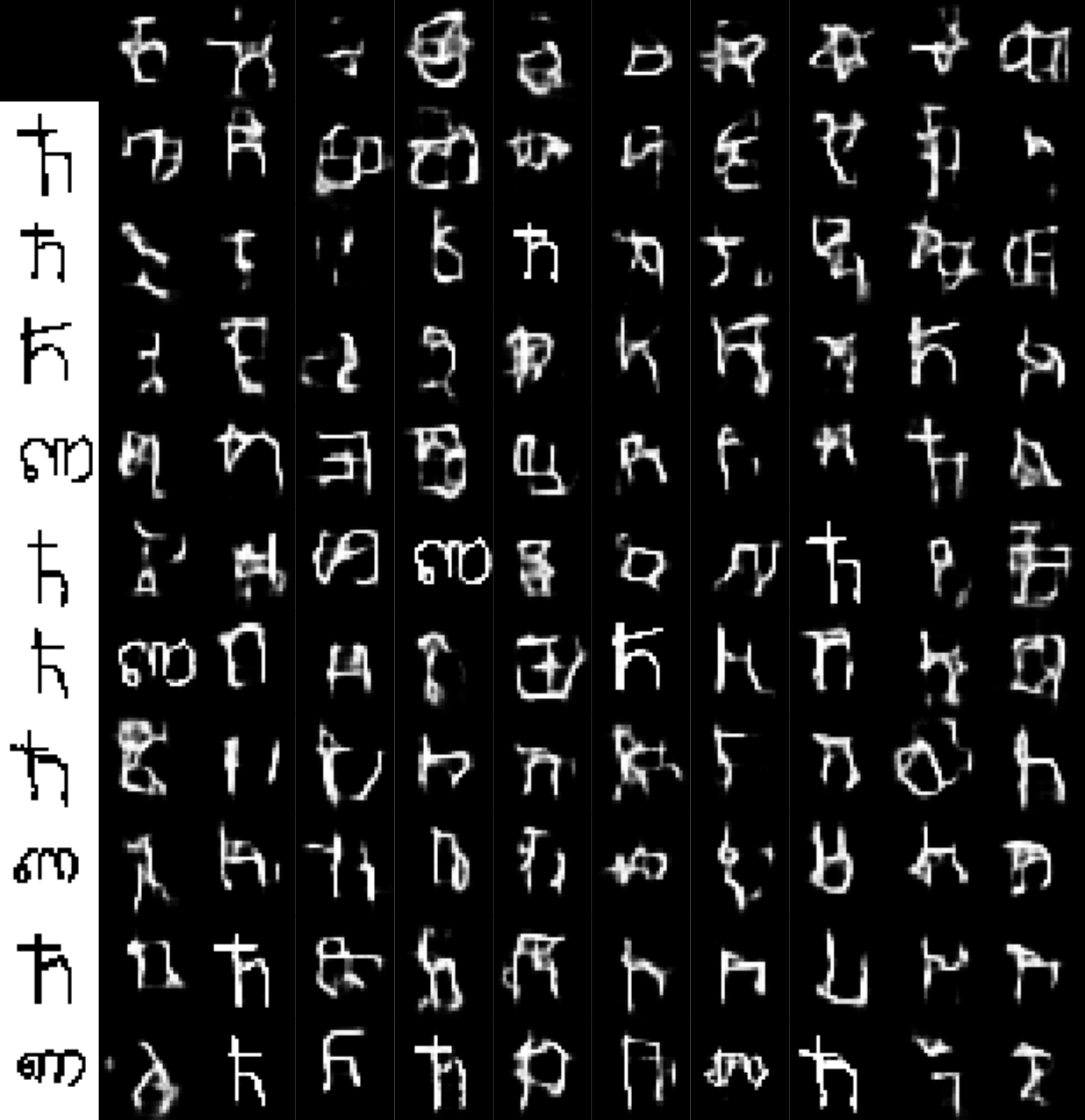} & \includegraphics[width=2.5in]{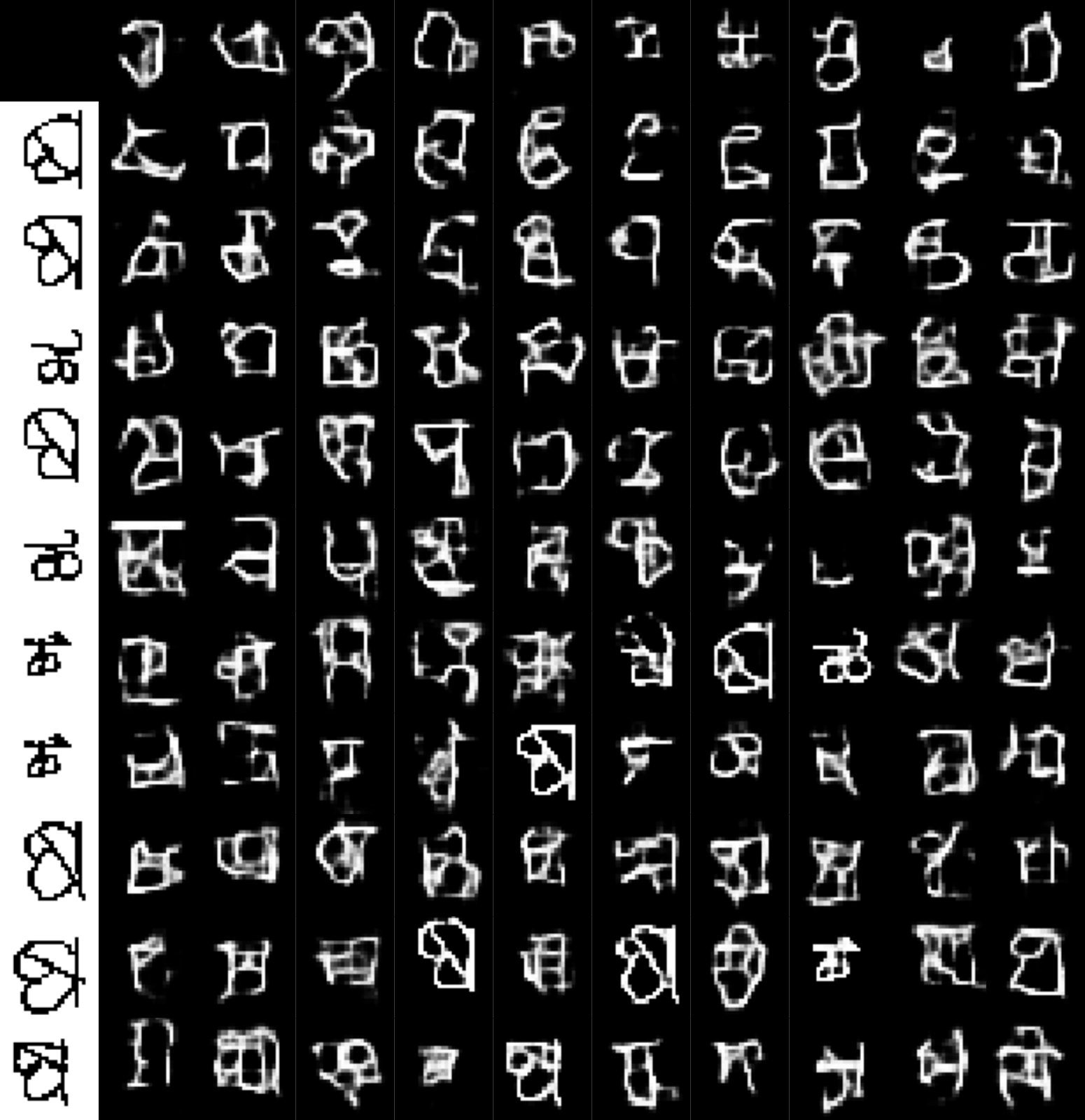} \\
\end{tabular}
\caption{Additional conditionally-generated samples from the GMN, one pseudo-input used}
\label{fig:additional_samples_gmn}
\end{figure*}

We provide more samples from the model on figures~\ref{fig:additional_samples_gmn_conditional} and~\ref{fig:additional_samples_gmn}.

\section{Data-dependent prior}\label{sec:prior}

We adapt the full matching procedure used in the generative and recognition models to define a data-dependent prior.
As before, we use the hidden state $\hstate$ of a recurrent controller to match the conditioning data:
\begin{equation}\label{eq:prior_matching}
    a_P(\hstate_k, \obsvars_t) = \frac{\exp(\text{sim}(f_P(\hstate_k), g_P(\obsvars_t, \hstate_k)))}{\sum_{t'=1}^T \exp(\text{sim}(f_P(\hstate_k), g_P(\obsvars_{t'}, \hstate_k)))},
\end{equation}
and then aggregate the result:
\begin{equation}\label{eq:prior_prototype}
    \result_P^{k+1} = \sum_{t=1}^T a_P(\hstate_k, \obsvars_t) \psi_P(\obsvars_t, \hstate_k), \quad \hstate_{k+1} = R(\hstate_k, \result^k_P),
\end{equation}
where subscript $\cdot_P$ denotes ``prior''.

As opposed to the conditional likelihood and the recognition model, only the hidden state is passed to function $f_P$ as there is no other information to use for matching.

The prior is then parametrized as a normal distribution with diagonal covariance: $p(\hidvars | \Obsvars, \params) = \mathcal{N}(\hidvars | \mu(\result^k_P, \hstate_{K+1}), \Sigma(\result^k_P, \hstate_{K+1}))$.

\section{Model architecture}

In order to reduce the number of trainable parameters, we shared some functionality across different parts of our model.
In particular, we set functions $f_G = f_R$ to be equal and also completely shared prototype extractors: $\psi_G = \psi_R = \psi_P$.
As we mentioned, the generative part and the recognition model also shared the recurrent controller used in the full context matching procedure, but the prior had its own controller.
All controllers were implemented as GRU~\citep{chung-icml15-gated} maintaining $200$-dimensional hidden state.

Feature extractors $g_\cdot$ were also identical.
Each function $f$ or $g$ used in our model is simply an affine transformation of feature encoder's output (and a hiddent state of a recurrent controller in the case of full context matching) to a $200$-dimensional space followed by a non-linearity.

By default, we used a parametric rectified linear function as a non-linearity everywhere where applicable.

\subsection{Conditional generator}

The conditional generator network producing parameters for $p(\obsvars | \hidvars, \Obsvars, \params)$ has concatenation of $\hidvars$ and the output of the matching operation $[\result, \hstate]$ as input which is transformed to $3 \times 3 \times 32$ tensor and then passed through $3$ residual blocks of transposed convolutions.
Each block has the following form:
\begin{align*}
    h &= \text{conv}_1(x), \\
    y &= f(\text{conv}_2(h) + h) + \text{pool}(\text{scale}(x)),
\end{align*}
where $f$ is a non-linearity which in our architecture is always parametric rectified linear function~\citep{he2015delving}.

The block is parametrized by size of filters used in convolutions $\text{conv}_1$ and $\text{conv}_2$, shared number of filters $F$ and stride $S$.
\begin{itemize}
    \item $\text{scale}$ is another convolution with $1 \times 1$ filters and the shared stride $S$.
    \item In all other convolutions number of filters is the same and equals $F$.
    \item $\text{conv}_1$ and $\text{pool}$ have also stride $S$.
    \item $\text{conv}_2$ preserve size of the input by padding and has stride $1$.
\end{itemize}

Blocks used in our paper have the following parameters $(W_1 \times H_1, W_2 \times H_2, F, S)$:
\begin{enumerate}
    \item $(2 \times 2, 2 \times 2, 32, 2)$
    \item $(3 \times 3, 3 \times 3, 16, 2)$
    \item $(4 \times 4, 3 \times 3, 16, 2)$
\end{enumerate}

Then log-probabilities for binary pixels were obtained by summing the result of these convolutions along the channel dimension.

\subsection{Feature encoder $\psi$}

Function $\psi$ has an architecture which is symmetric from the generator network.
The only difference is that the $\text{scale}$ scale operation is replaced by bilinear upscaling. 

The residual blocks for feature encoder has following parameters:
\begin{enumerate}
    \item $(4 \times 4, 3 \times 3, 16, 2)$
    \item $(3 \times 3, 3 \times 3, 16, 2)$
    \item $(2 \times 2, 2 \times 2, 32, 2)$
\end{enumerate}

The result is a tensor of $3 \times 3 \times 32 = 288$ dimensions.

\section{Effect of number of attention steps}\label{sec:num_steps}

\begin{figure*}
    \centering \includegraphics[width=\textwidth]{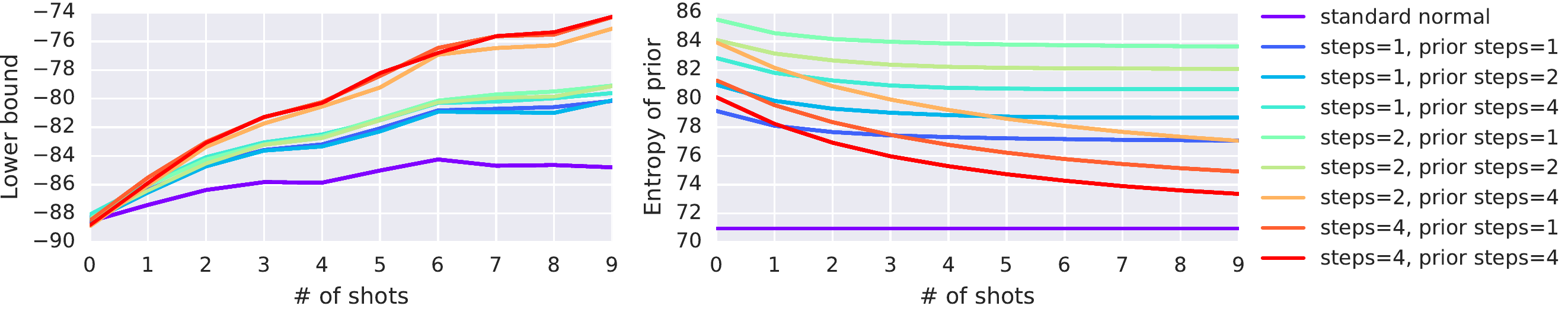}
    \caption{Lower bound estimates (left) and entropy of prior (right) for various numbers of attention steps and numbers of conditioning examples. Numbers are reported for the training part of Omniglot.}\label{fig:steps}
\end{figure*}

Since the full context matching procedure consists of multiple attention steps, it is interesting to see the effect of these numbers on model's performance.
We trained several models with smaller architecture and $T=10$ varying number of attention steps allowed for the likelihood and recognition shared controller and the prior controller respectively. 
The models were compared using exponential moving averages of lower bounds corresponding to different numbers of conditioning examples $\Obsvars_{<t}$ obtained during the training.
Results of the comparison can be found on figure~\ref{fig:steps}.

Interestingly, larger numbers of steps lead to better results, however lower bounds are almost not improving after the shared controller is allowed for $4$ steps. 
This behaviour was not observed with discriminative matching networks perhaps confirming the difficulty of unsupervised learning.
Another important result is that the standard Gaussian prior makes adaptation significantly harder for the model yet still possible which justifies importance of the data-dependent prior (see section~\ref{sec:prior}).

One may also see that all models preferred to set higher variances for a prior resulting to higher entropy comparing to standard normal prior.
Clearly as more examples are available, generative matching networks become more certain about the data and output less dispersed Gaussians.

Based on this study, for our experiments we have chosen a model with 4 attention steps in the shared controller and single step in the prior controller, which is a reasonable compromise between computational cost and performance.

\section{Transfer to MNIST}

In this experiment we test the ability of generative matching networks to adapt not just to new concepts, but also to a new \emph{domain}.
Since we trained our models on $28 \times 28$ resolution for Omniglot it should be possible to apply them on MNIST dataset as well.
We used the test part of MNIST to which we applied a single random binarization.

Table~\ref{tbl:mnist_predictive} contains estimated predictive likelihood for different models.
Qualitative results from the evaluation on Omniglot remain the same.
Although transfer to a new domain caused significant drop in performance for all of the models, one may see that generative matching networks still demonstrate the ability to adapt to conditioning data.
At the same time, average matching does not seem to efficiently re-use the conditioned data in such transfer task since relative improvements in expected conditional log-likelihood are rather small.
Apparently, the model trained on a one-class datasets also learned highly dataset-dependent features as it actually performed even worse than the model with $C_{\text{train}}=2$. 

\begin{figure*}
        \begin{subfigure}[t]{0.33\textwidth}
               \includegraphics[width=\linewidth]{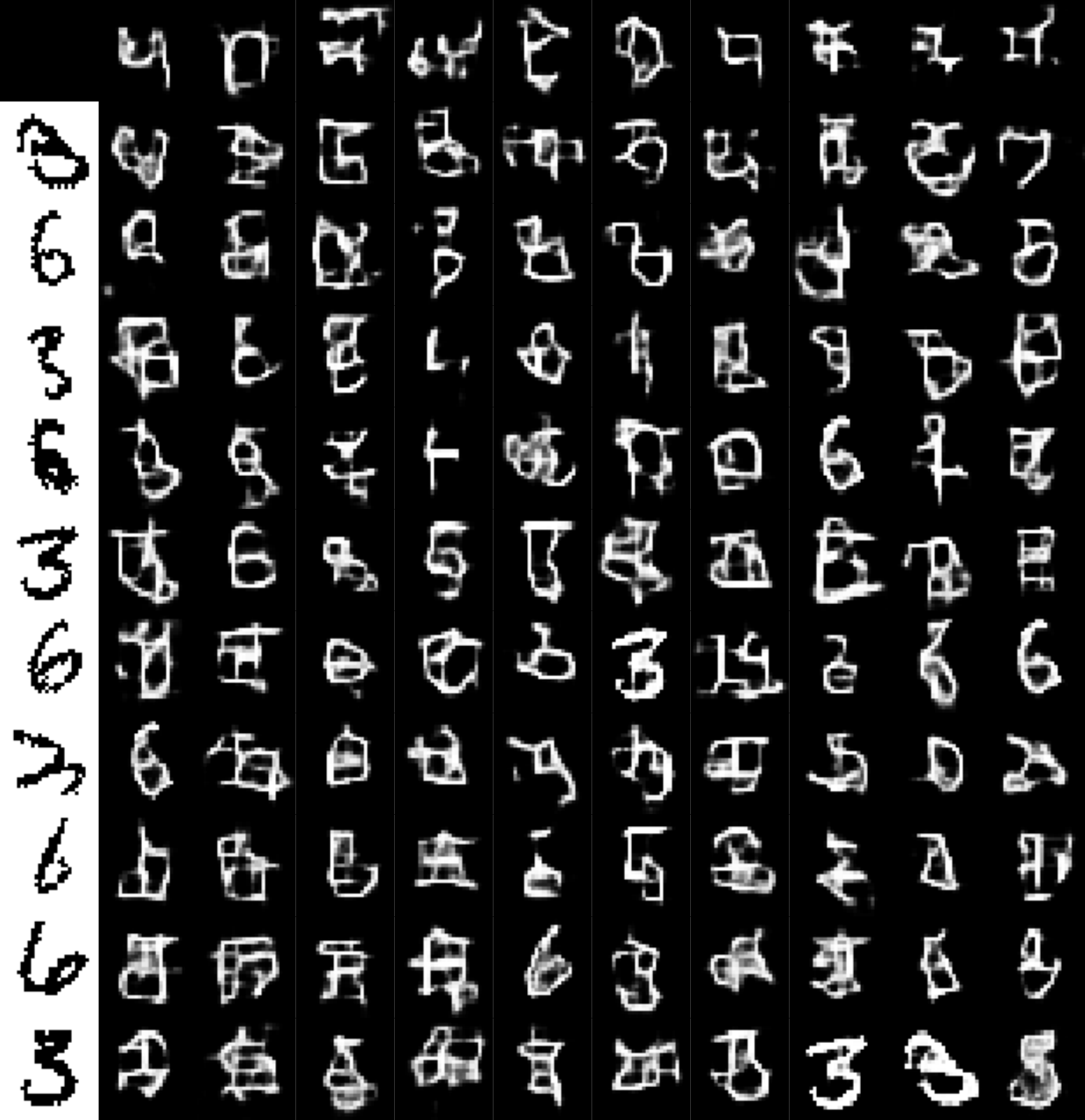}
                \caption{GMN}
                \label{fig:mnist_fce_pi}
        \end{subfigure}%
        \begin{subfigure}[t]{0.33\textwidth}
               \includegraphics[width=\linewidth]{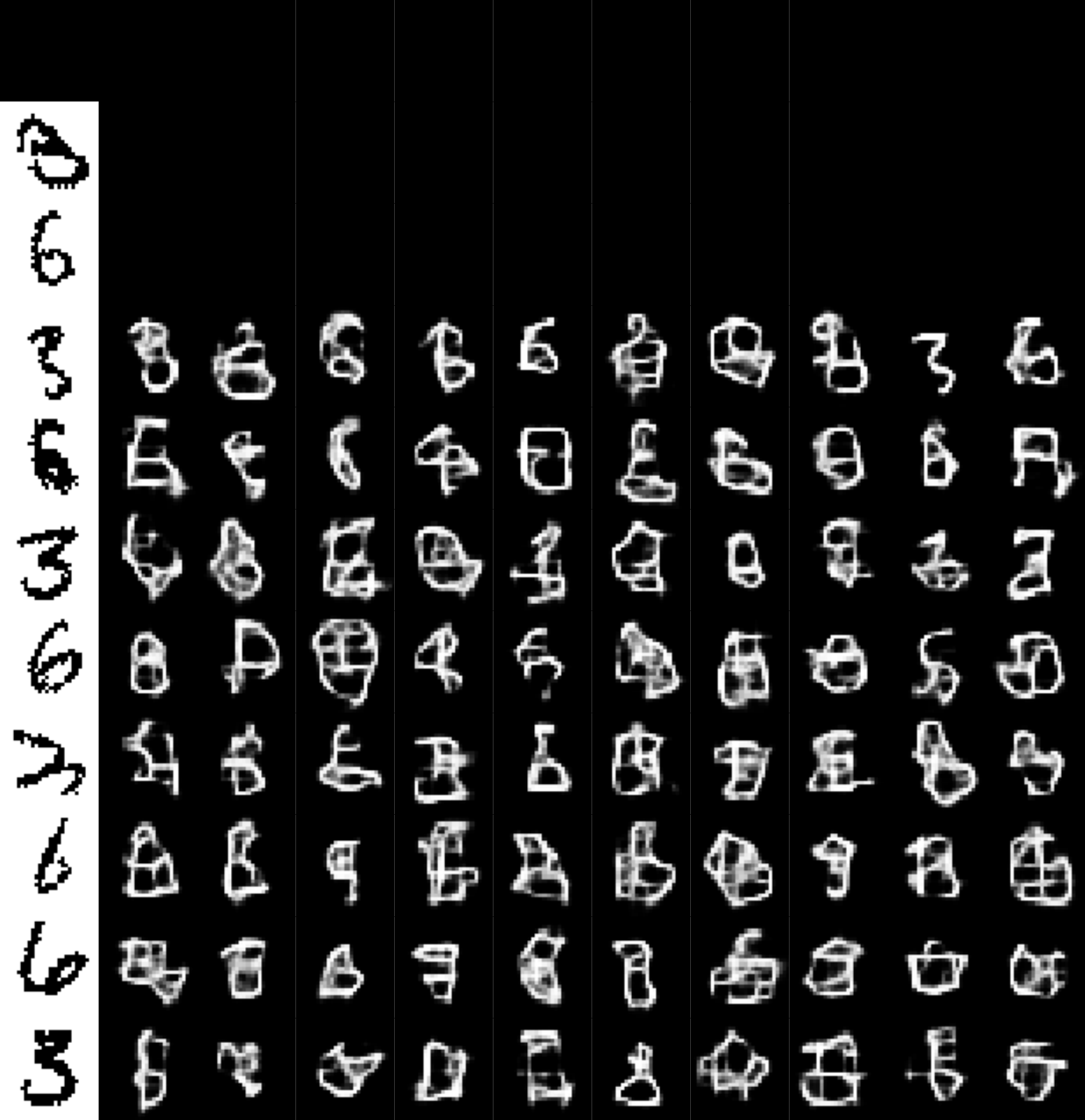}
                \caption{GMN, no pseudo-input}
                \label{fig:mnist_fce}
        \end{subfigure}%
        \begin{subfigure}[t]{0.33\textwidth}
               \includegraphics[width=\linewidth]{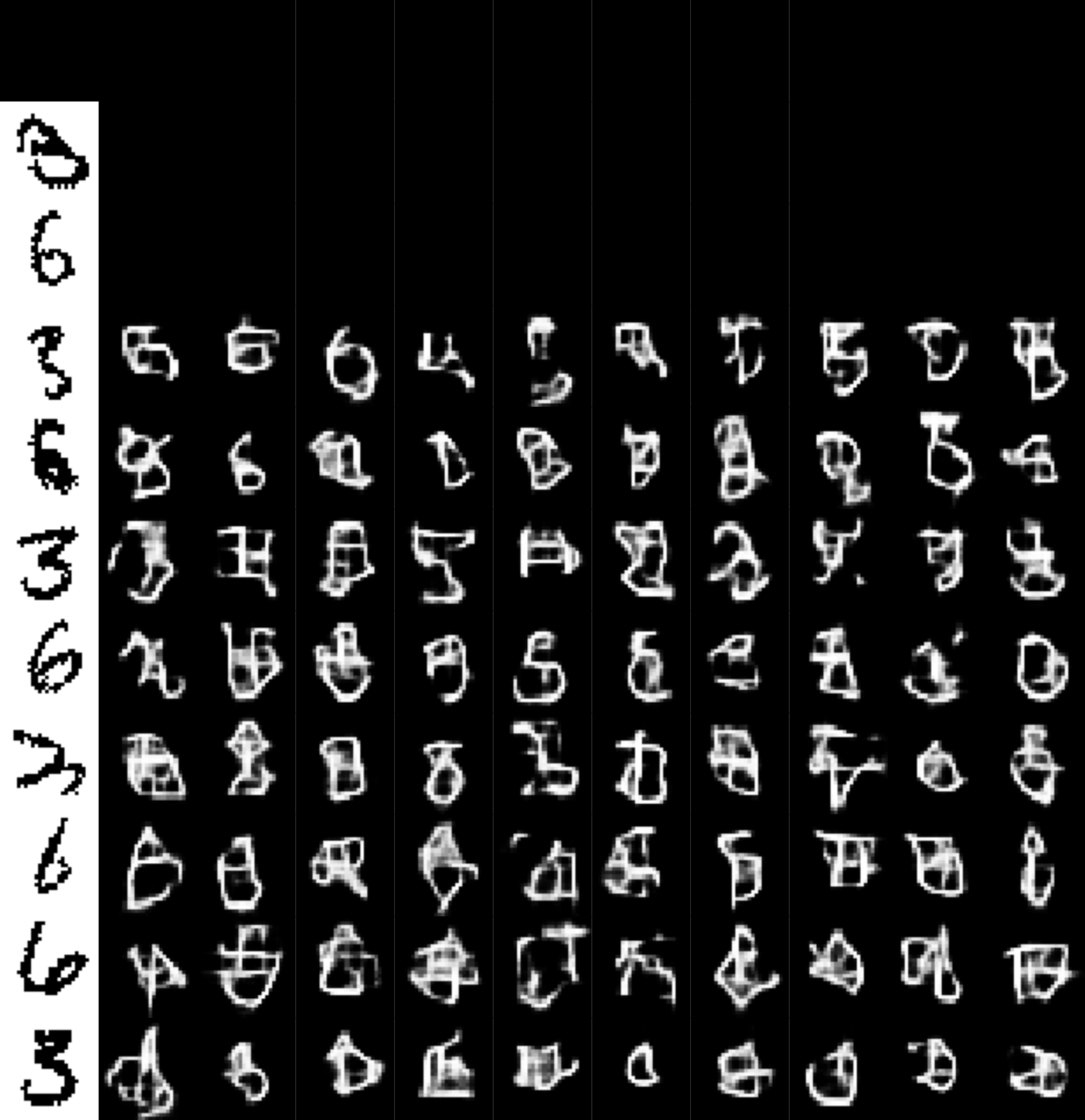}
                \caption{GMN, no attention, no pseudo-input}
                \label{fig:mnist_avg}
        \end{subfigure}
        \caption{Conditionally generated samples on MNIST. Models were trained on the train part of Omniglot. Format of the figure is similar to fig. 2 in the main paper. }\label{fig:mnist_samples}
\end{figure*}

We also provide conditional samples on figure~\ref{fig:mnist_samples}.
Both visual quality of samples and test log-likelihoods are significantly worse comparing to Omniglot which can be caused by a visual difference of the MNIST digits from Omniglot characters.
The images are bolder and less regular due to binarization.
\citet{edwards2016towards} suggest that the quality of transfer may be improved by augmentation of the training data, however for the sake of experimental simplicity and reproducibility we resisted from any augmentation.

\begin{table*}[t]
\setlength{\tabcolsep}{5pt}
\caption{Conditional negative log-likelihoods for the test part of MNIST. Models were trained on the train part of Omniglot.}
\label{tbl:mnist_predictive}
\centering
\begin{tabular}{lccccccccc}
\textbf{}                             & \multicolumn{1}{l}{\textbf{}} & \multicolumn{8}{c}{\textbf{Number of conditioning examples}}                                                            \\
\textbf{Model}                        & \textbf{$C_{\text{test}}$}               & \textbf{0} & \textbf{1} & \textbf{2} & \textbf{3} & \textbf{4} & \textbf{5} & \textbf{10} & \textbf{19} \\
\hline \\
GMN, $C_{\text{train}}=2$                   & 1                             & 126.7     & 121.1     & 118.4     & 117.6     & 117.1     & 117.1     & 117.1      & 118.5      \\
GMN, $C_{\text{train}}=2$                   & 2                             & 126.2     & 123.1     & 121.3     & 120.1     & 119.4     & 118.9     & 118.3      & 119.6      \\
GMN, $C_{\text{train}}=2$, no pseudo-input      & 1                             &            & 135.1    & 120.9     & 117.5     & 115.7     & 114.4     & 111.7      & 109.8      \\
GMN, $C_{\text{train}}=2$, no pseudo-input      & 2                             &            &            & 123.1     & 121.9     & 119.4     & 118.8     & 115.2      & 113.2      \\
\hline
GMN, $C_{\text{train}}=1$, avg              & 1                             & 131.5     & 126.5     & 123.3     & 121.9     & 121.0     & 120.2     & 118.6      & 117.5      \\
GMN, $C_{\text{train}}=2$, avg              & 2                             & 126.2     & 122.8     & 121.0     & 119.9     & 118.9     & 118.7    & 117.8      & 116.8      \\
GMN, $C_{\text{train}}=1$, avg, no pseudo-input & 1                             &            & 132.1     & 126.9     & 125.0     & 124.8     & 123.9     & 121.7      & 120.9      \\
GMN, $C_{\text{train}}=2$, avg, no pseudo-input & 2                             &            &            & 118.4     & 117.9     & 117.4     & 117.1     & 116.6      & 115.8      \\    
\end{tabular}
\end{table*}

\section{Evaluation of the neural statistician model}

The neural statistician model falls into the category of models with global latent variables which we describe in section 2.2.
The conditional likelihood for this model has the following form:
\begin{equation}\label{eq:ns_predictive}
    p(\obsvars | \Obsvars) = \int p(\globvars | \Obsvars) \int p(\hidvars | \globvars) p(\obsvars | \globvars, \hidvars) d \hidvars d \globvars.
\end{equation}

This quantity is hard to compute since it consists of an expectation with respect to the true posterior over global variable $\globvars$.
In our evaluation we have tried three different strategies for computing the above-mentioned integral.

First, we used self-normalizing importance sampling to directly estimate $p(\obsvars | \Obsvars, \params)$ as 
\begin{equation*}
\hat{p}(\obsvars | \Obsvars, \params) = \frac{\sum_{s=1}^S w_s p(\obsvars, \hidvars^{(s)} | \globvars^{(s)}, \params)}{\sum_{s=1}^S w_s}, 
w_s = \frac{p(\globvars^{(s)}, \Obsvars, \Hidvars^{(s)} | \params)}{q(\globvars^{(s)} | \Obsvars, \varparams) q(\Hidvars^{(s)}, \hidvars^{(s)} | \Obsvars, \obsvars, \globvars^{(s)}, \varparams)},
\end{equation*}
with samples $ \globvars^{(s)} $ and $ \hidvars^{(s)} $ obtained from the recognition model. 
We observed somewhat contradictory results such as non-monotonic dependency of the estimate on the size of conditioning dataset.
The diagnostic of the effective sample size suggested that the recognition model is not well suited as proposal for the task.

Another strategy was to sequentially estimate $p(\Obsvars_{<t}, \params)$ and then use the equation 
$$p(\obsvars_t | \Obsvars_{<t}, \params) = \frac{p(\obsvars_{t}, \Obsvars_{<t} | \params)}{p(\Obsvars_{<t} | \params)},$$
which appeared to as unreliable as the previous strategy.

Finally, we decided to use the approximate posterior $q(\globvars | \Obsvars)$ in equation~\eqref{eq:ns_predictive} that was learned together with the model instead of the exact one.
Practically, one can almost never access the true posterior and when using the model would rather resort to the recognition model.
Hence, the resulting approximate predictive distribution and the corresponding estimate is aligned with practical usage of the model and is often considered in the literature~\citep{snelson2005compact,jaakkola2000bayesian}.
Fortunately, the approximate posterior is easy to sample from by construction, being a multivariate Normal distribution.
The final estimate was computed as:
\begin{equation*}
\hat{p}(\obsvars | \Obsvars, \params) = \sum_{s=1}^S \frac{ w_s p(\obsvars, \hidvars^{(s)} | \globvars^{(s)}, \params)}{S}, \quad
w_s = \frac{p(\hidvars^{(s)} | \globvars^{(s)}, \params)}{q( \hidvars^{(s)} | \obsvars, \globvars^{(s)}, \varparams)}, 
\end{equation*}
where samples are, again, obtained from the recognition model.

\end{document}